\documentclass[nohyperref]{article}

\usepackage{microtype}
\usepackage{graphicx}
\usepackage{subfigure}
\usepackage{booktabs} % for professional tables
\usepackage{multirow}
\usepackage[table,xcdraw]{xcolor}
\usepackage{soul}
\usepackage{float}

\usepackage{hyperref}

\usepackage[accepted]{icml2022}

\usepackage{amsmath}
\usepackage{amssymb}
\usepackage{mathtools}
\usepackage{amsthm}

\usepackage[capitalize,noabbrev]{cleveref}

\usepackage{appendix}

\theoremstyle{plain}

\theoremstyle{definition}

\theoremstyle{remark}

\usepackage[textsize=tiny]{todonotes}

\usepackage{xcolor}
\usepackage{nidanfloat}

\icmltitlerunning{Interpretable Hidden Markov Model-Based Deep Reinforcement Learning}

\begin{document}

\twocolumn[
\icmltitle{Interpretable Hidden Markov Model-Based Deep Reinforcement Learning Hierarchical Framework for Predictive Maintenance of Turbofan Engines}




\begin{icmlauthorlist}
\icmlauthor{Ammar N. Abbas}{comp}
\icmlauthor{Georgios Chasparis}{comp}
\icmlauthor{John D. Kelleher}{yyy}
\end{icmlauthorlist}

\icmlaffiliation{yyy}{Department of Computer Science, Technological University of Dublin, Dublin, Ireland}
\icmlaffiliation{comp}{Software Competence Center Hagenberg, Hagenberg, Austria}

\icmlcorrespondingauthor{Ammar Abbas}{ammar.abbas@scch.at}

\icmlkeywords{Deep Reinforcement Learning (DRL), industrial maintenance, Hidden Markov Model (HMM), decision-making under uncertainty, and interpretable AI}

\vskip 0.3in
]



\printAffiliationsAndNotice{} % otherwise use the standard text.

\begin{abstract}
An open research question in deep reinforcement learning is how to focus the policy learning of key decisions within a sparse domain. This paper emphasizes combining the advantages of input-output hidden Markov models and reinforcement learning towards interpretable maintenance decisions. We propose a novel hierarchical-modeling methodology that, at a high level, detects and interprets the root cause of a failure as well as the health degradation of the turbofan engine, while, at a low level, it provides the optimal replacement policy. It outperforms the baseline performance of deep reinforcement learning methods applied directly to the raw data or when using a hidden Markov model without such a specialized hierarchy. It also provides comparable performance to prior work, however, with the additional benefit of interpretability.
\end{abstract}

%***************************************************************************************************************************************
%***************************************************************************************************************************************

\section{Introduction}
\label{introduction}
Machine learning has the potential to improve the performance of  equipment maintenance systems by providing accurate predictions regarding the type of equipment that should be replaced as well as the optimal replacement time. Predictive maintenance can be categorized as (i) \textit{Prognosis}: predicting failure and notifying for replacement or repair ahead of time  (\textit{Remaining Useful Life} (RUL) is usually used as a prognosis approach, which is the estimation of the remaining life of equipment or a system at any point in time before which it is no longer in a functional state \cite{sikorska2011prognostic}); (ii) \textit{Diagnosis}: predicting the actual cause of failure in the future through cause-effect analysis, or (iii) \textit{Proactive Maintenance}: anticipate and mitigate the failure modes and conditions before they develop within a certain equipment. \cite{d15}. In this paper, the aforementioned questions will be investigated in the context of predictive maintenance of turbofan engines \cite{sg08, c21}.

Reinforcement Learning (RL) is a natural approach to solving time-series-based stochastic decision problems, such as predictive maintenance \cite{sm20} and recently, it has shown promising results. RL systems learn by interacting with the environment and can learn in an online setting without having the predefined dataset beforehand; incorporating stochastic events \cite{sb18}. However, when the key policy decision learned by an RL agent is relatively rare in a dataset (such as the decision of when to change the equipment before failure while maximizing the use of each piece of equipment) the policy can be dominated by irrelevant phenomena in the data, resulting in inefficient training. At the same time, the derived optimal policy does not provide interpretations or the root cause of the failure, and therefore, it keeps humans out of the loop with limited collaborative intelligence. Furthermore, in real-world industrial environments, RL learns directly from the observed raw sensor data that does not provide information about the unobserved hidden factors responsible for the decision-making of the system such as its health, which limits the agent to behave sub-optimally. 

Hidden Markov Model (HMM) \cite{1165342} can overcome the challenges faced by RL through (i) learning unobserved states and interpretation based on those hidden states, (ii) combining multiple sensor data and leveraging on the covariance between the data, defining the state of the system and its hierarchical distribution, and (iii) dimensionality reduction based on the number of latent states that defines the model and reduces the size and complexity of the raw data \cite{yoon_lee_hovakimyan_2019}. In order to address the need for a more direct and specialized data-based optimization, while maintaining the interpretability of the derived policies, we propose an unsupervised hierarchical modeling technique that combines a high-level input-output hidden Markov model (IOHMM) with a low-level deep RL methodology for predictive maintenance. \textit{Hierarchical Reinforcement Learning (HRL)} is a solution towards sample efficient RL, which decomposes the long-horizon enormous state space into several short-horizon specialized tasks. At a first step, the IOHMM prefilters large amounts of non-relevant data generated during the normal running of the equipment and detects the state at which failure is imminent. At a second step, the deep RL agent learns a policy on equipment replacement conditioned on these (close to failure) states. Our experimental results indicate that the proposed state-/event-based approach with dynamic data pre-filtering has comparable performance to prior work that trains RL agents directly on the full dataset, hence increasing the training efficiency. Lastly, it allows for more explicit interpretability of the derived policies by learning the latent state spaces. Specifically, the IOHMM learns the hidden state representation of the system ($x_t$) and the DRL constructs the state-action pair modeling of the environment ($s_t,a_t$).

\textbf{Structure:}
\cref{sec:lit-rev} provides the literature review on hierarchical and event-driven RL as well as applications of DRL and HMM in interpretable industrial maintenance along with this paper's contributions. \cref{sec:use-case} provides the overview of the use case. \cref{sec:pm-as-rl} frames predictive maintenance as an RL problem. \cref{sec:prop_meth} proposes the novel methodology. \cref{sec:exp-set} explains the experimental setup and baseline architectures. \cref{sec:exp1} provides the interpretability aspect of the proposed methodology. Finally, \cref{sec:exp2} compares the proposed architecture with baseline and prior work.

%***************************************************************************************************************************************
%***************************************************************************************************************************************

\section{Related Work}
\label{sec:lit-rev}
The major challenge for HRL is the ability to learn the hierarchical structure that requires a priori knowledge or supervision from experts as discussed by \cite{pateria2021hierarchical, yu2018towards}. References \cite{xu2021interpretable, lyu2019sdrl} mention data inefficiency and lack of interpretability as the major challenges of RL, hence a new hierarchical RL framework is proposed by the authors that uses symbolic RL. similarly, \cite{lee2021attaining} solves the same aforementioned problem through HRL by decomposing states into pretrained primitives. The effectiveness of an adaptive event-driven RL strategy and its convergence proof is shown in \cite{meng2019adaptive}. Further, \cite{parra2021event} proposes an event-driven explainable RL methodology.

Multi-Objective Reinforcement Learning (MORL) is used by \cite{l20} as a predictive maintenance strategy in the steel industry, where the model learns from both its own experience through environment interaction as well as from the human experience feedback using a policy-shaping approach \cite{g13}. Double Deep Q-Learning (DDQN) approach \cite{vgs16} for developing a general-purpose predictive maintenance architecture is used by \cite{ony20}. They highlight the difference between a traditional regression model and a self-learning agent that provides the recommended actions for each piece of equipment in the system. Authors have used turbofan engines \cite{sg08} as their case study and have discussed the limitations of prior work that just estimate the RUL of a system, giving no cause-effect relationship between the failure and the components of the equipment. Instead of using raw sensor data, they have dimensionally reduced it into one principal component indicating the Health Index (HI) of the equipment. Using the same case study, \cite{sm20}, provide an optimal maintenance decision and RUL prediction-based alarm system at any predefined cycle before failure, using Bayesian Network-based deep RL. \textit{Bayesian particle filtering}; a Bayesian approach based on sequential Monte Carlo simulation \cite{chen2003bayesian} is used on top of DRL to map the raw sensor data into latent belief degradation states.

The use of HMM is proposed by \cite{g11} for predicting RUL of turbofan engines that demonstrated its effectiveness towards the interpretation of the fault point with a sudden decrease in RUL and transition of HMM state. Similarly, \cite{hofmann2020hidden} uses HMM for predicting a failure event by using hierarchical mixtures of distributions to predict the overall failure rate and degradation path through individual assets. Input-Output Hidden Markov Model (IOHMM) \cite{bf95} is explored by \cite{7984302} for failure diagnosis, prognosis, and health state monitoring of a diesel generator in an online setting. Similarly, \cite{shahin2019estimating} have shown how IOHMM can be used for prognosis and diagnosis by predicting the hidden (health) state and RUL through a simulated example. The effectiveness of online HMM estimation-based Q learning that converges to a higher mean reward for \textit{Partially Observable Markov Decision Process (POMDP)}, where certain variables are hidden and not directly observable is proved mathematically by \cite{yoon2019hidden}.

\textbf{Literature Gap and Research Contributions:}
The majority of the research in predictive maintenance using RL is focused on prognosis based on the RUL estimation from multivariate raw sensor readings. However, the interpretability of the faults of the machine (at the equipment level) is missing. Furthermore, realistic environments often have partial observability, where learning from raw data might lead to suboptimal decisions. Additionally, RL encounters learning inefficiency when trained with limited samples and in an online setting \cite{dulac2021challenges}. In this paper, a novel methodology for maintenance decisions and interpretability is proposed that is based on a hierarchical DRL. At a high level, an IOHMM is designed for detecting imminent-to-failure states, while at a low level, a DRL is designed for optimizing the optimal replacement policy. We further present an extensive comparative analysis with prior work that demonstrates the effectiveness of the proposed methodology in terms of both performance and interpretation.
%***************************************************************************************************************************************

\section{Use Case: Turbofan Engines}
\label{sec:use-case}

NASA Commercial Modular Aero-Propulsion System Simulation (C-MAPSS), turbofan engine degradation dataset \cite{sg08} is widely used in the community of predictive maintenance. The dataset consists of several engine units with multivariate time-series sensor readings and operating conditions discretized based on the flight cycles. Each unit observes some initial degradation at the start of the equipment failure, after which the health of the equipment degrades exponentially until it reaches a final failure state, hence, having a run-to-failure simulation. However, these degrees of wear are unknown. Recently, NASA published an updated version of the dataset \cite{c21} that records the real-time flight data and appends the operational history to the degradation modeling. This dataset additionally provides the ground truth values for the health state of the engine based on the component failure modes. These subsets of the datasets will be used in this paper: \linebreak
\textbf{FD001} with 1 operating condition and 1 failure mode.\linebreak
\textbf{FD002} with 6 operating conditions and 1 failure mode.\linebreak
\textbf{FD003} with 1 operating condition and 2 failure modes.\linebreak
\textbf{DS01 (version 2)} with ground truth values of degradation. \linebreak

%***************************************************************************************************************************************
%***************************************************************************************************************************************
\section{Framing Predictive Maintenance as a Reinforcement Learning Problem}
\label{sec:pm-as-rl}

In this section, the decision-making problem associated with optimal predictive maintenance is framed as an RL problem. A general modeling technique is described here followed by its simplified version for our use case.

%***************************************************************************************************************************************
\subsection{Environment Dynamics and Modeling}
\label{env-dyn}

\begin{figure}[ht]
\vskip 0.2in
\begin{center}
\centerline{\includegraphics[width=\columnwidth]{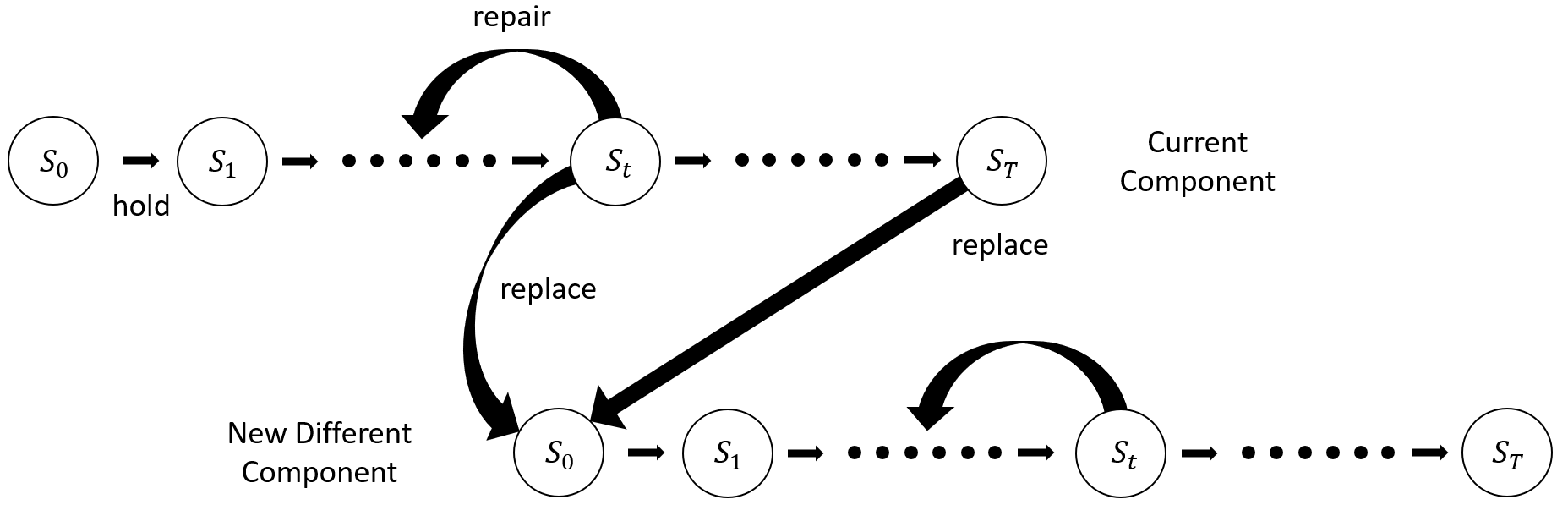}}
\caption{Environment model dynamics}
\label{fig:env_mod}
\end{center}
\vskip -0.2in
\end{figure}

\cite{ony20} consider three actions as a general methodology for any decision-making maintenance model; hold, repair, and replace. The constraints can be the maintenance budget and the objective function can be the maximum uptime of the equipment. We propose a general framework for modeling such environments with state transitions based on the actions selected under stochasticity (uncertainty of failure, and randomness of replacement by new equipment) at any state, as illustrated in \cref{fig:env_mod}. An action to hold transitions to the next state in time, under uncertainty of ending up in a failure state. An action to repair transitions the current state of the equipment back in its life cycle to an arbitrary state as defined by the type of repair or through some standards either from experience, reference manual, or history of data. An action to replace the equipment taken at the current state, transitions it to the initial state of the next equipment (introducing randomness), however, if the equipment reaches its final (failure) state regardless of the action chosen; the equipment must be replaced now. \cref{alg:env_mod} of \cref{append-algo} defines the modeling of such stochastic dynamics in terms of RL used within the Open AI gym environment \cite{b16}. For simplicity and lack of data for repair actions, the action space consists of just two actions (hold or replace).

%***************************************************************************************************************************************

\subsection{Reward Formulation}
The reward signal is one of the main factors responsible for an effective RL, therefore, it should be chosen carefully. For the maintenance decision in a simplified case, having just the replacement or hold actions, a dynamic reward structure has been formulated as shown in \cref{eq-reward} \cite{sm20}. This cost formulation maintains the trade-off between the early replacement ($c_r$) and replacement after failure ($c_r + c_f$).

\begin{equation}
\label{eq-reward}
r_{t}= \begin{cases}0, & \mathrm{a}_{t}=\text { Hold } $\quad\quad\& $ \mathrm{t}<T_{j}, \\ -\frac{c_{r}}{t}, & \mathrm{a}_{t}=\text { Replace} $\quad\& $  \mathrm{t}<T_{j}, \\ -\frac{c_{r}+c_{f}}{T_{j}}, & \mathrm{a}_{t}=\text { Hold } $\quad\quad\& $ \mathrm{t}=T_{j}, \\ -\frac{c_{r}+c_{f}}{T_{j}}, & \mathrm{a}_{t}=\text { Replace} $\quad\& $ \mathrm{t}=T_{j} .\end{cases}
\end{equation}

%***************************************************************************************************************************************

\subsection{Evaluation Criteria}
\label{eval}
To evaluate the performance of the RL agent, numerical values were chosen for better comparison. 

%***************************************************************************************************************************************
\subsubsection{Cost}
The average optimal total return/cost ($\widetilde{Q^{*}}$) serves as a single numeric value used and compared with the upper and lower bounds of cost possible for such conditions \cite{sm20}. 

\paragraph{Ideal Maintenance Cost (IMC)}
serves as the lower bound and the ideal cost in such maintenance applications. It is the incurred cost when the replacement action is performed one cycle before the failure, utilizing its maximum potential with the minimum cost as shown in \cref{eq:IMC}. 
\begin{equation}
\label{eq:IMC}
\phi_{IMC} \approx \frac{N \cdot c_{r}}{N \cdot(\mathbb{E}(T)-1)} \approx \frac{N \cdot c_{r}}{\sum_{j=1}^{N}\left(T_{j}-1\right)}
\end{equation}

\paragraph{Corrective Maintenance Cost (CMC)}
serves as the upper bound and the maximum cost in such maintenance applications. It is the incurred cost when the replacement action is performed after the equipment has failed as shown in \cref{eq:CMC}. 
\begin{equation}
\label{eq:CMC}
\phi_{CMC} \approx \frac{\left(c_{r}+c_{f}\right)}{\mathbb{E}(T)} \approx \frac{N \cdot\left(c_{r}+c_{f}\right)}{\sum_{j=1}^{N} T_{j}}
\end{equation}

\paragraph{Average Optimal Cost ($\widetilde{Q^{*}}$)}
is the average cost that the agent receives as its performance on the test set as shown in \cref{eq:avg-q}. 
\begin{equation}
\label{eq:avg-q}
\widetilde{Q^{*}}(s, a)=\frac{1}{N} \sum\left[r(s, a)+\gamma \max _{a^{\prime}} Q^{*}\left(s^{\prime}, a^{\prime}\right)\right]
\end{equation}

\subsubsection{Average useful life before replacement}
It quantifies; how many useful life cycles on average are remaining when the replacement action is proposed by the agent. Ideally, it should be 1 according to our defined optimization criteria.

%***************************************************************************************************************************************

\section{Proposed Methodology}
\label{sec:prop_meth}

\begin{figure}[ht]
\vskip 0.2in
\begin{center}
\centerline{\includegraphics[width=0.9\columnwidth]{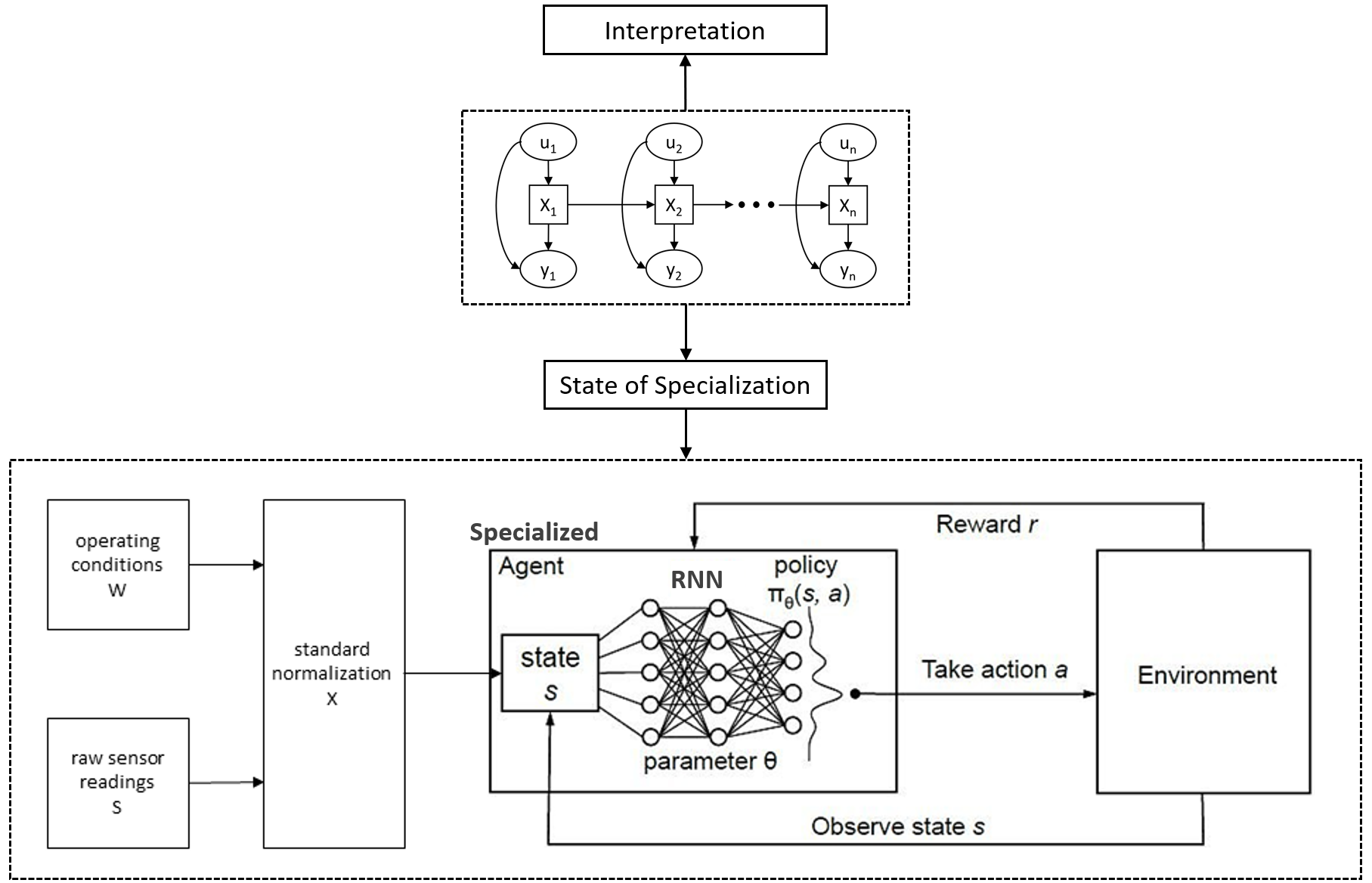}}
\caption{Specialized Reinforcement Learning Agent (SRLA).}
\label{fig:srla}
\end{center}
\vskip -0.2in
\end{figure}

The proposed methodology is a hierarchical model integrating an IOHMM and DRL agent. Within this hierarchical model, the purpose of the IOHMM is to identify when the system is approaching a desired (in our case: failure) state. Once the IOHMM has entered this failure state, the task of the DRL agent is to optimize the decision of when to replace the equipment to maximize its total useful life. This IOHMM-DRL model introduces hierarchical levels into the information space and allows for the state- or event-based optimization. This further allows for a more efficient DRL training, since the training dataset is restricted to the imminent-to-failure states. Such agents can be deployed under situation-dependent adaptations as mentioned in \cite{pb21}. Beyond the performance considerations of the model, the IOHMM component provides a level of interpretability in terms of identifying failure states (leading towards RUL estimation), root cause of failure, and health degradation stages. \cref{fig:srla} illustrates the proposed hierarchical model which we name Specialized Reinforcement Learning Agent (SRLA). 

The DRL training and optimation process is relatively standard. We use Deep Learning (DL) as a function approximator that generalizes effectively to enormous state-action spaces through the approximation of unvisited states \cite{bertsekas1996neuro} as shown in \cref{eq:drl}.

\begin{equation}
\label{eq:drl}
\begin{aligned}
& L_{i}\left(\theta_{i}\right)=\mathbb{E}_{a \sim \mu}\left[\left(y_{i}-Q\left(s, a ; \theta_{i}\right)\right)^{2}\right]; \\
& y_{i}:=\mathbb{E}_{a^{\prime} \sim \pi}\left[r+\gamma \max _{a^{\prime}} Q\left(s^{\prime}, a^{\prime} ; \theta_{i-1}\right) \mid\right. \left.S_{t}=s, A_{t}=a\right]
\end{aligned}
\end{equation}

At a high level, an IOHMM is employed that is an extension to a standard HMM model \cite{bf95}. In a standard HMM model (as described \cite{rabiner1989tutorial}) the training optimization objective is to identify the model parameters that best determine the given sequence of observations. To predict the probability of being in a particular hidden state, given the observation sequence $Y$ and trained model parameters $\lambda$ (initial state, transition, and emission probability matrices), Equation~\ref{eq:gamma} is used. $\gamma$ is the vector defining the probability of being in each hidden state at a particular time, which will be used as the input to DRL in our baseline extension. Equation~\ref{eq:maxprob_q} predicts the most probable hidden state that in this context leads to the health degradation state given the sequence of sensor observations. However, this does not provide the information of the most probable sequence of states; as it might be possible that the two most probable states at a particular time step may not be the most optimal state sequence. This problem is solved by the Viterbi algorithm \cite{forney1973viterbi} as shown in Equation~\ref{eq:delta}, where, in this context, $\delta$ is used to predict the health degradation sequence, where the last cycle of each equipment determines the failure state.

\begin{eqnarray}
\label{eq:gamma}
\gamma_{t}(i) & = & P\left(x_{t}=S_{i} \mid Y, \lambda\right) \\
\label{eq:maxprob_q}
x_{t} & = & \underset{1 \leq i \leq N}{\operatorname{argmax}}\left[\gamma_{t}(i)\right], \quad 1 \leq t \leq T \\
\label{eq:delta}
\delta_{t}(i) & = & \max _{x_{1}, \cdots, x_{t-1}} P\left[x_{1} \cdots x_{t}=i, Y_{1} \cdots Y_{t} \mid \lambda\right]
\end{eqnarray}

One of the limitations of HMM is that the mathematical model does not take into account any input conditions that affect the state transition and the emission probability distribution of the observations (outputs). In the context of industrial settings, these inputs are the operating conditions that heavily influence the state of the system and control the system's behavior. Therefore, IOHMM is used to have a more general model architecture that can utilize the information of operating conditions, which modifies Equation~(\ref{eq:gamma} and \ref{eq:delta}) to Equation~(\ref{eq:gamma-mod} and \ref{eq:delta-mod}) with $\lambda$ being conditioned on the input ($U$) as well.

\begin{eqnarray}
\label{eq:gamma-mod}
\gamma_{t}(i) & = & P\left(x_{t}=S_{i} \mid U, Y, \lambda\right) \\
\label{eq:delta-mod}
\delta_{t}(i) & = & \max _{x_{1}, \cdots, x_{t-1}} P\left[x_{1} \cdots x_{t}=i, Y_{1} \cdots Y_{t} \mid U, \lambda\right] \notag \\
\end{eqnarray}

%***************************************************************************************************************************************

\section{Experimental Setup}
\label{sec:exp-set}
The baseline systems defined in this paper are distinguished and designed by varying each of these four stages: (i) input, (ii) feature engineering, (iii) RL architecture, and (iv) output. System 1 just uses the raw sensor data. System 2 signifies the importance of using raw sensor data along with operating conditions. It is used to set the cost of failure to be used for the rest of the experiments. System 3 uses HMM as the layer between the raw sensor readings and the DRL. Its significance is to (i) determine the optimal number of HMM states to be used in the experiments, and (ii) demonstrate the effectiveness of HMM and compare it with system 1. Finally, System 4 uses IOHMM in place of HMM to introduce a more general architecture that can take into account varying operating conditions. Implementation of HMM and IOHMM is done through libraries \cite{hmmlearn, iohmm}. The summary of the 
training parameters are shown in \cref{app-sec:train_param} of \cref{append-algo}. 

%***************************************************************************************************************************************

\paragraph{System 1: Baseline\\} 
(i) Raw sensor data as the input, (ii) Standard normalization as the feature extraction module, (iii) DNN as the RL architecture, and (iv) Action policy at the output.

\paragraph{System 2: Baseline + Operating Conditions\\} 
(i) Raw sensor data and operating conditions as the input, (ii) Standard normalization as the feature extraction module, (iii) DNN as the RL architecture, and (iv) Action policy at the output.

\begin{figure}[ht]
\vskip 0.2in
\begin{center}
\centerline{\includegraphics[width=\columnwidth]{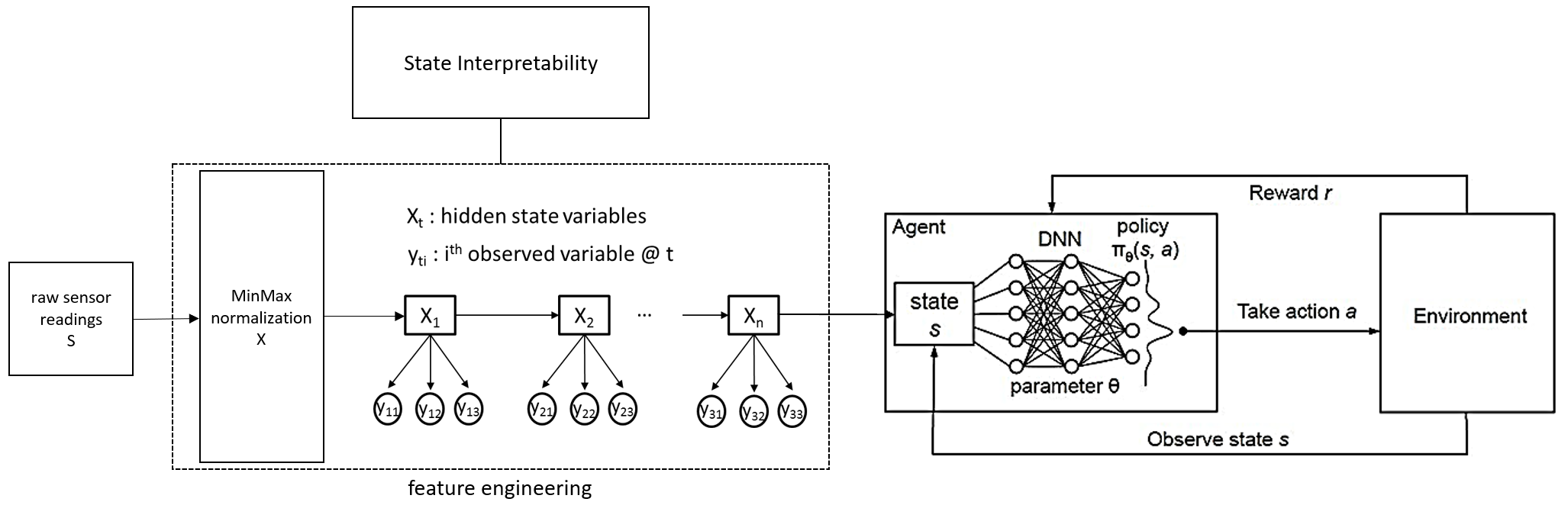}}
\caption{HMM posterior probabilities as the input to DRL.}
\label{methodology3}
\end{center}
\vskip -0.2in
\end{figure}

\paragraph{System 3: Baseline + HMM\\} 
(i) Raw sensor data as the input, (ii) MinMax normalization and HMM as the feature extraction module, (iii) DNN as the RL architecture, and (iv) Action policy, RUL estimation, and event-based unsupervised clustering and interpretation at the output; as shown in \cref{methodology3}.

\begin{figure}[ht]
\vskip 0.2in
\begin{center}
\centerline{\includegraphics[width=\columnwidth]{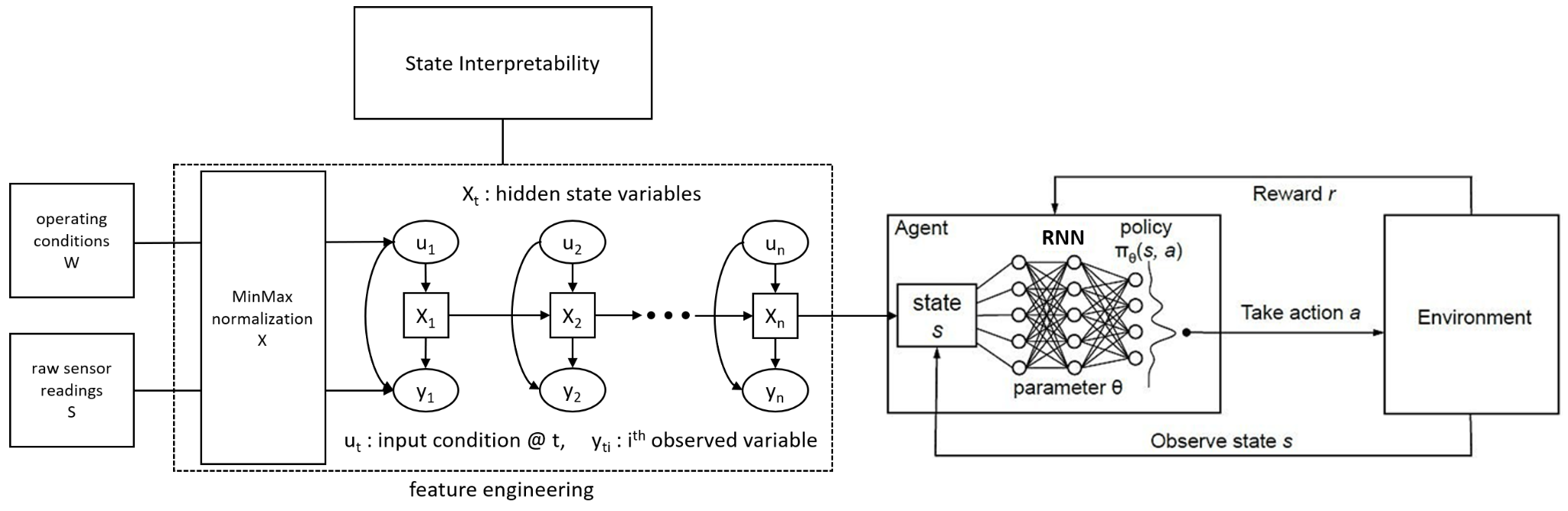}}
\caption{IOHMM posterior probabilities as the input to DRL.}
\label{methodology4}
\end{center}
\vskip -0.2in
\end{figure}

\paragraph{System 4: Baseline + Operating Conditions + IOHMM\\}
\label{sec:sys4}
(i) Raw sensor data and operating conditions as the input, (ii) MinMax normalization and IOHMM as the feature engineering module, (iii) RNN as the RL architecture, and (iv) Action policy, RUL estimation, and event-based unsupervised clustering and interpretation at the output; as shown in \cref{methodology4}. 

%***************************************************************************************************************************************
%***************************************************************************************************************************************

\subsection{Model's Hyperparameters Search}
\label{sec:hyp-sear}
This section consists of determining the hyperparameters (i) cost of failure ($c_f$) and (ii) HMM states. While searching for these parameters, we also observe the effectiveness of HMM and address the question of how well DRL can perform by learning through hidden states? Hence, the effectiveness of the architectures has been evaluated as described in \cref{eval}. The dataset used for this part of the experiment is FD001, which is split into an 80:20 (train:test) ratio. 

%***************************************************************************************************************************************

\subsubsection{Calculating the cost of failure}
The reward function (Equation~\ref{eq-reward}) for RL agent assumes a \emph{cost of failure} ($c_f$) and \emph{cost of replacement} ($c_r$) to be specified. However, the NASA C-MAPSS dataset does not specify this parameter. To fix a value for this, we train System 2 using a range of different $c_f$, while fixing $c_r$ and then comparing and identifying the $c_f$ that minimizes the average of total optimal cost per episode ($\widetilde{Q^{*}}$). We used System 2 because this system has the baseline architecture while at the same time using the full set of input parameters available in the dataset. $c_r$ is fixed (100) and the comparison is based on the different $c_f$ (25, 500, and 1000) as shown in \cref{tab:result_eval}. It was observed that as the $c_f$ increases, $\widetilde{Q^{*}}$ gets closer to the ideal cost as well as the number of failed units decreases to 0\%. However, the agent becomes more cautious suggesting replacement action earlier in the lifetime of the engine; thereby, increasing the average remaining cycles. Therefore, a balance between the dynamic replacement cost (decreasing cost with increased life cycles) and the higher failure cost must be maintained. \cref{tab:result_eval} also shows the results of the optimal action policy learned by the agent through System 1. As a comparison, it can be concluded that the additional information of the operating conditions helps the model to learn a better maintenance policy.

%***************************************************************************************************************************************

\subsubsection{Calculating the number of hidden states}
System 3 has been used here to find the number of states of HMM model that maximizes the likelihood for our state space as well as the performance of DRL through an iterative process. 15 states of the HMM showed better performance results than the rest. The model trained through HMM gives the posterior probability distribution for every state as shown in Equation~\ref{eq:gamma}, which is then fed as an input to the DRL agent to be able to learn the optimal maintenance (replacement) policy. The experiment was performed on the test set using the failure cost of 1000 and with the same parameters as the previous experiment for better comparison. The model with the HMM outperforms System 1 and System 2 as shown in \cref{tab:result_eval}.

\begin{table}[t]
\caption{Comparative evaluation and hyperparameter search.}
\label{tab:result_eval}
\vskip 0.15in
\begin{center}
\begin{small}
\begin{sc}
\begin{tabular}{c
>{\columncolor[HTML]{D9D9D9}}c 
>{\columncolor[HTML]{D9D9D9}}c 
>{\columncolor[HTML]{D9D9D9}}c cc}
\hline
\begin{tabular}[c]{@{}c@{}}Fail\\ cost\end{tabular} & \begin{tabular}[c]{@{}c@{}}Avg\\ Q*\end{tabular} & IMC & CMC & \begin{tabular}[c]{@{}c@{}}Average\\ remaining\\ cycles\end{tabular} & \begin{tabular}[c]{@{}c@{}}Failed \\ units\end{tabular} \\ \hline
\multicolumn{6}{c}{System 2}                                                                  \\ \hline
25 & 0.54 & 0.45 & 0.56 & 2.4 & 45\% \\
500 & 0.61 & 0.45 & 2.68 & 7.5 & 5\% \\
\cellcolor[HTML]{D9EAD3}1000 & \cellcolor[HTML]{B6D7A8}0.49 & \cellcolor[HTML]{B6D7A8}0.45 & \cellcolor[HTML]{B6D7A8}4.92 & \cellcolor[HTML]{D9EAD3}7.0 & \cellcolor[HTML]{D9EAD3}0\% 
\\ \hline
\multicolumn{6}{c}{System 1} 
\\ \hline
1000 & 0.51 & 0.45 & 4.92 & 12.0 & 0\% \\ 
\hline
\multicolumn{6}{c}{System 3} 
\\ \hline
\begin{tabular}[c]{@{}c@{}}HMM\\states\end{tabular} & & & & &
\\ \hline
5 & 0.60 & 0.45 & 4.92 & 44.8 & 0\% \\
10 & 0.54 & 0.45 & 4.92 & 24.2 & 0\% \\
\cellcolor[HTML]{D9EAD3}15 & \cellcolor[HTML]{B6D7A8}0.49 & \cellcolor[HTML]{B6D7A8}0.45 & \cellcolor[HTML]{B6D7A8}4.92 & \cellcolor[HTML]{D9EAD3}6.8 & \cellcolor[HTML]{D9EAD3}0\% \\
20 & 0.53 & 0.45 & 4.92 & 20.2 & 0\% \\
30 & 0.55 & 0.45 & 4.92 & 28.5 & 0\% 
\\ \hline
\end{tabular}
\end{sc}
\end{small}
\end{center}
\vskip -0.1in
\end{table}

%***************************************************************************************************************************************
%***************************************************************************************************************************************
%***************************************************************************************************************************************

\section{Experiment 1: Interpretations Based on the Hidden States}
\label{sec:exp1}
Datasets FD003 and DS01 are used in this section with System 3 for event-based hypothesis and state interpretations. The experiments performed here are to address the question, can the hidden state help towards interpretability?

%***************************************************************************************************************************************
%***************************************************************************************************************************************

\subsection{Interpretability - Failure Event Hypothesis}
\label{sec:int-fail}
\begin{figure}[ht]
\vskip 0.2in
\begin{center}
    \subfigure[]{\includegraphics[width=0.87\columnwidth]{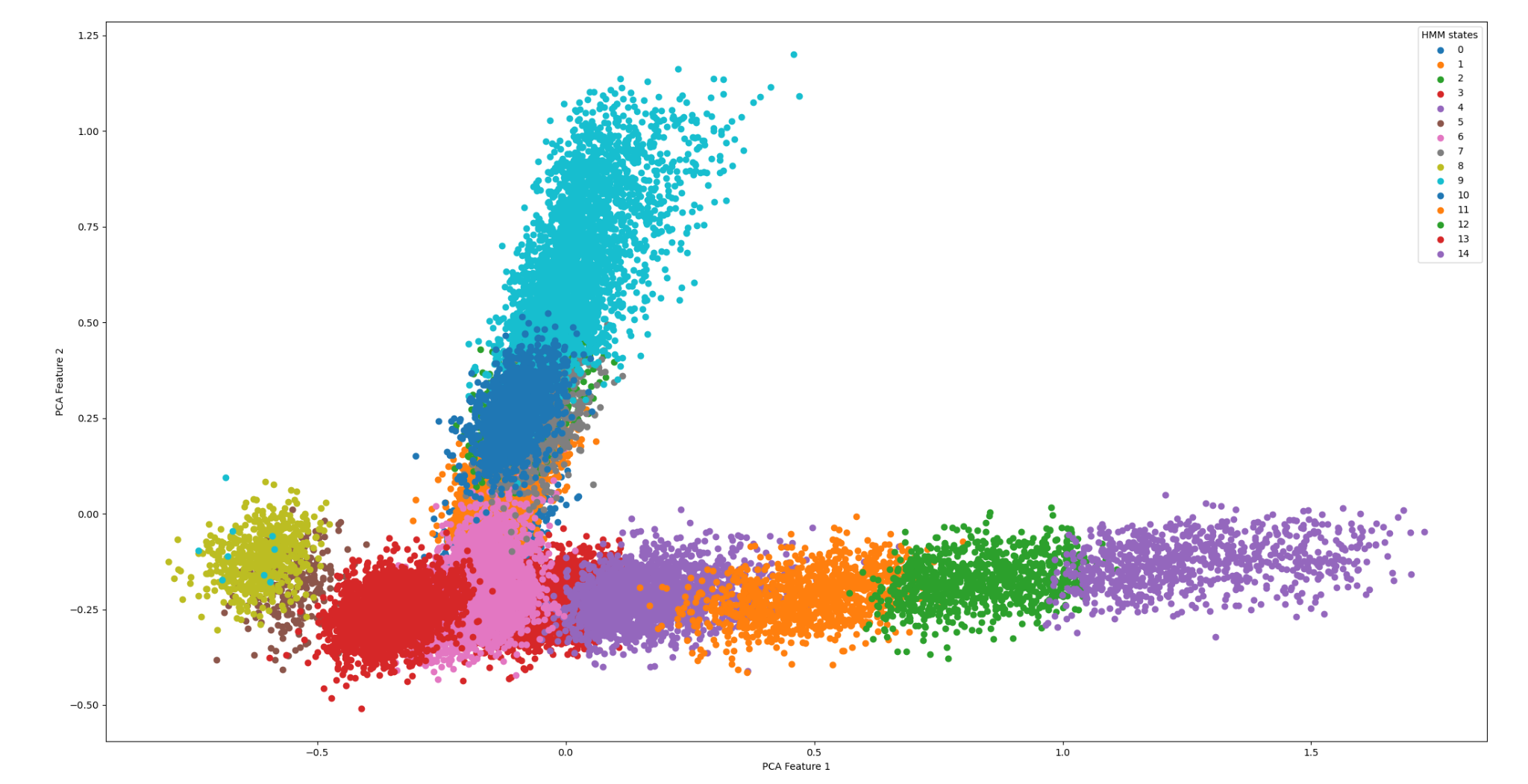}}\quad
    \subfigure[]{\includegraphics[width=0.87\columnwidth]{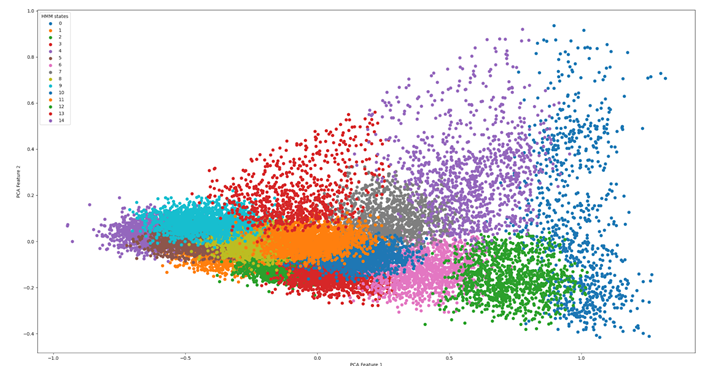}}\quad
\caption{HMM states clustering for (a) FD003 and (b) FD001.}
\label{fig:hmm_clust}
\end{center}
\vskip -0.2in
\end{figure}

Based on the state sequence distributions predicted by the HMM through Viterbi algorithm from Equation~\ref{eq:delta}, each state of a particular event can be decoded, such as the failure mode or the degradation stage as shown in \cite{g11}. Due to the unavailability of the ground truth for other state mappings in FD003, just the failure states (last cycle state) were mapped in this experiment. \cref{fig:hmm_clust}(a) plots state distributions for each data point based on the hidden states of the HMM on FD003. The data points are collected from the sensor readings of every engine per cycle; reduced to 2D features through Principal Component Analysis (PCA) for visualization. It can be hypothesized that each of these state clusters defines a particular event. The dataset was provided with the insight that it consists of 2 failure modes (HPC and fan degradation), however, the ground truth for the engines corresponding to which failure mode was not provided. Analyzing the failure states revealed two states that corresponded to the failure event (state 9 and 14) as visualized in \cref{fig:state_decoding-fd003} of \cref{append:fig}, which might be based on the two failure modes. Just to validate this hypothesis it was compared with FD001 as shown in \cref{fig:hmm_clust}(b), where only one failure mode exists as defined in the dataset description and further analysis showed that only one state (state 0) was observed to be the failure state for each engine as shown in \cref{fig:state_decoding-fd001} of \cref{append:fig}. Therefore, giving an initial heuristics of HMM state distribution based on the failure events.

To discover the most relevant sensor readings corresponding to these failure states that triggered the HMM to predict such a state, feature importance was performed. Raw sensor readings were used as the input feature to the model and HMM state predictions based on the Viterbi algorithm were used as the target. After fitting the model, the importance of each sensor could be extracted for each HMM state. \cref{tab:feat_imp} shows the subset of the output of the feature importance for the failed states. The features with relatively higher score were selected from each class and its corresponding actual sensor information and description were extracted from \cite{sg08} as described in \cref{tab:feat_sensor}. From the background information of the sensor descriptions, it was observed that the sensor importance for two different states showed a concrete failure event interpretation that corresponded to the failure described in the dataset (HPC and Fan degradation), as hypothesized in \cref{tab:sensor_event}. \cref{alg:feat-imp} of \cref{append-algo} defines the feature importance usage in the context of HMM state interpretation.

\begin{table}[t]
\caption{Feature (sensor) Importance.}
\label{tab:feat_imp}
\vskip 0.15in
\begin{center}
\begin{small}
\begin{sc}
\begin{tabular}{ll}
\hline
\multicolumn{1}{c}{State: 9} & \multicolumn{1}{c}{State: 14} \\ \hline
$feature\:5$: -12.497 & \cellcolor[HTML]{C0C0C0}$feature\:5$: 4.211 \\
$feature\:6$: -3.873 & $feature\:6$: 0.268 \\
$feature\:7$: -5.984 & $feature\:7$: 0.175 \\
$feature\:8$: 0.463 & \cellcolor[HTML]{C0C0C0}$feature\:8$: 19.697 \\
$feature\:9$: -7.529 & $feature\:9$: 0.325 \\
$feature\:10$: -12.737 & \cellcolor[HTML]{C0C0C0}$feature\:10$: 3.973 \\
$feature\:11$: -3.454 & $feature\:11$: 0.153 \\
$feature\:12$: -5.651 & $feature\:12$: 0.097 \\
\cellcolor[HTML]{C0C0C0}$feature\:13$: 4.036 & $feature\:13$: -3.555 \\ 
\hline
\end{tabular}
\end{sc}
\end{small}
\end{center}
\vskip -0.1in
\end{table}

\begin{table}[t]
\caption{Feature to sensor description.}
\label{tab:feat_sensor}
\vskip 0.15in
\begin{center}
\begin{small}
\begin{sc}
\begin{tabular}{lcc}
\hline
\multicolumn{1}{c}{Feature} & Sensor & Description \\ \hline
$feature\:5$ & $P_{30}$ & pressure at HPC outlet \\
$feature\:8$ & $epr$ & Engine pressure ratio \\
$feature\:10$ & $phi$ & fuel flow : pressure (HPC) \\
$feature\:13$ & $BPR$ & Bypass Ratio \\ \hline
\end{tabular}
\end{sc}
\end{small}
\end{center}
\vskip -0.1in
\end{table}

\begin{table}[t]
\caption{Sensor importance to failure event hypothesis.}
\label{tab:sensor_event}
\vskip 0.15in
\begin{center}
\begin{small}
\begin{sc}
\begin{tabular}{ccc}
\hline
\begin{tabular}[c]{@{}c@{}} HMM \\ state \end{tabular} & \begin{tabular}[c]{@{}c@{}}Important \\ sensor \\ reading\\\end{tabular} & \begin{tabular}[c]{@{}c@{}}Failure event \\ hypothesis\\(interpretation)\end{tabular} \\ \hline
9 & $BPR$ & Fan degradation \\
14 & $P_{30}$, $epr$, $phi$ & HPC degradation \\ \hline
\end{tabular}
\end{sc}
\end{small}
\end{center}
\vskip -0.1in
\end{table}

%***************************************************************************************************************************************

\subsubsection{Remaining Useful Life (RUL) estimation}
\begin{figure}[ht]
\vskip 0.2in
\begin{center}
\centerline{\includegraphics[width=0.85\columnwidth]{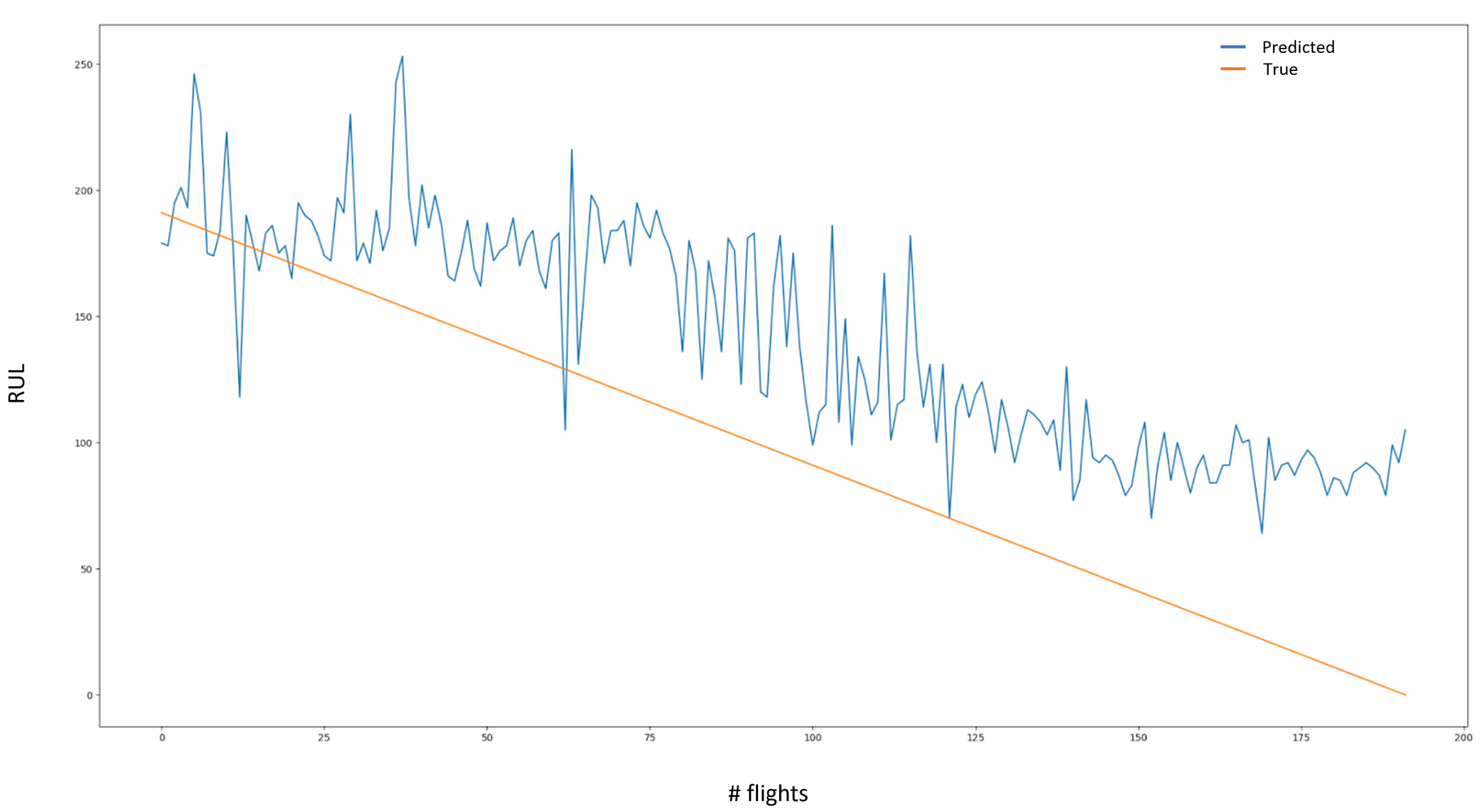}}
\caption{Remaining Useful Life estimation}
\label{fig:rul}
\end{center}
\vskip -0.2in
\end{figure}

Another additional benefit of using the HMMs is RUL prediction per every cycle in an unsupervised and online setting. The Viterbi algorithm was used to predict the optimal sequence of states until the cycle at a particular time step. Based on the last observed state, it can predict the next most probable state using the transition probability matrix. Using the emission probabilities, one can sample the sensor observations based on the predicted next state and append it to the previously seen observations sequence. This process was continued until the sequence predicted the next state to be the failure state as decoded through the methodology discussed in the earlier \cref{sec:int-fail}. The total number of transitions to the failure state gives the RUL at that particular cycle as elaborated in \cref{alg:rul_est} of \cref{append-algo}. For each cycle, the trend can be predicted as shown in \cref{fig:rul}.

%***************************************************************************************************************************************
%***************************************************************************************************************************************

\subsection{Interpretability - State Decoding and Mapping}
\label{sec:int-state}
\begin{figure}[ht]
\vskip 0.2in
    \begin{center}
      \subfigure[]{\includegraphics[width=0.475\columnwidth]{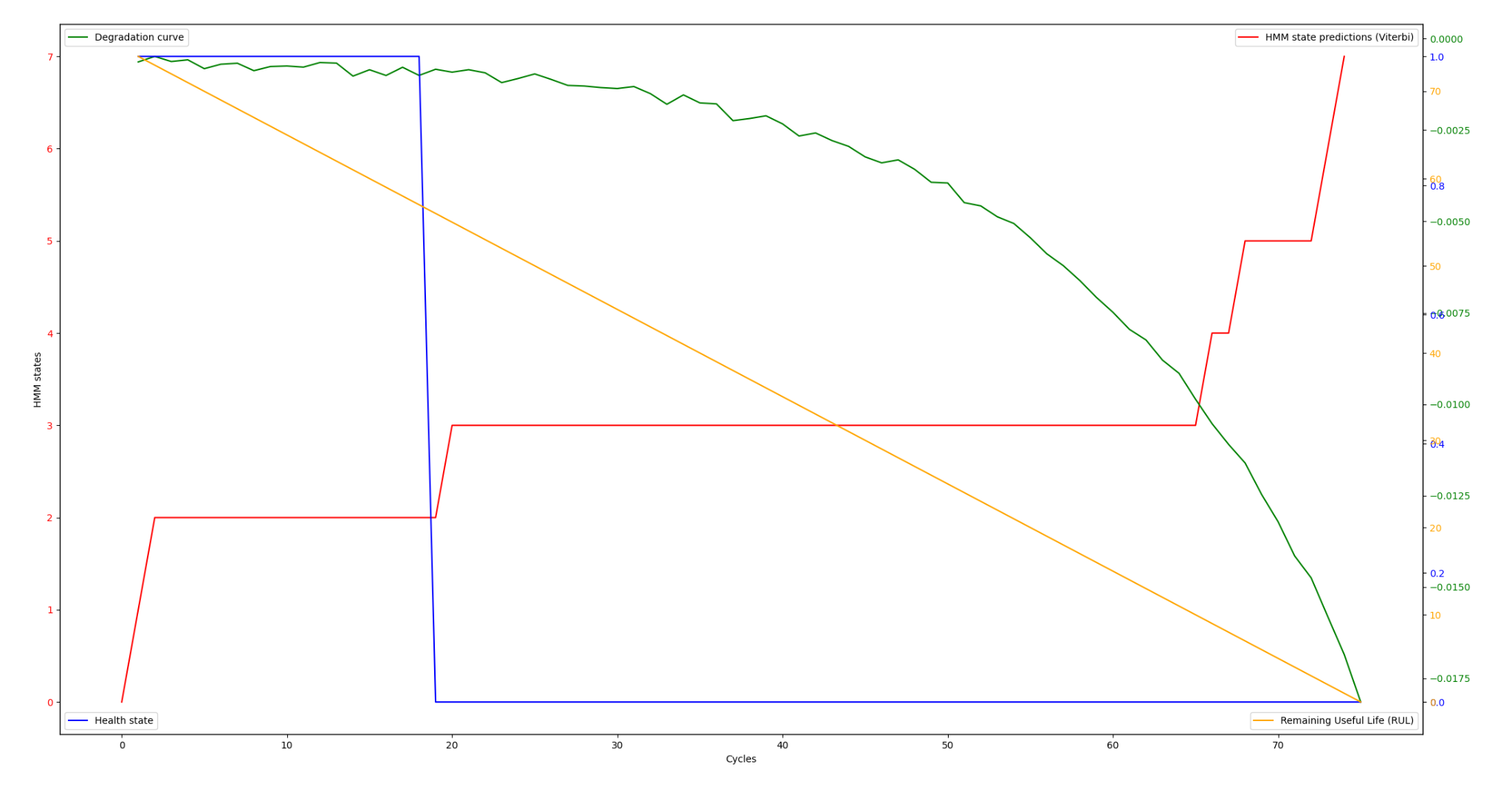}}\quad
      \subfigure[]{\includegraphics[width=0.475\columnwidth]{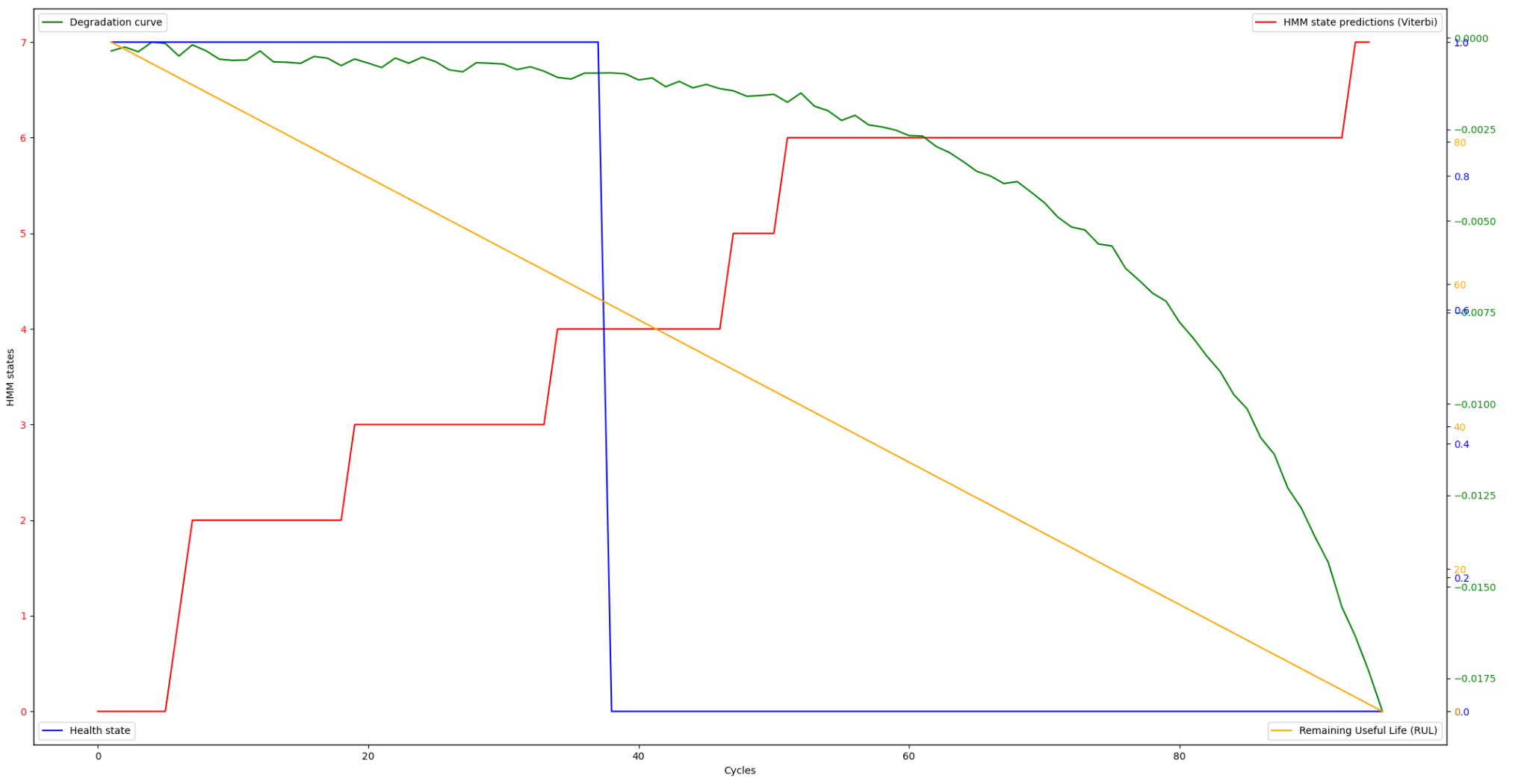}}\quad
      \subfigure[]{\includegraphics[width=0.475\columnwidth]{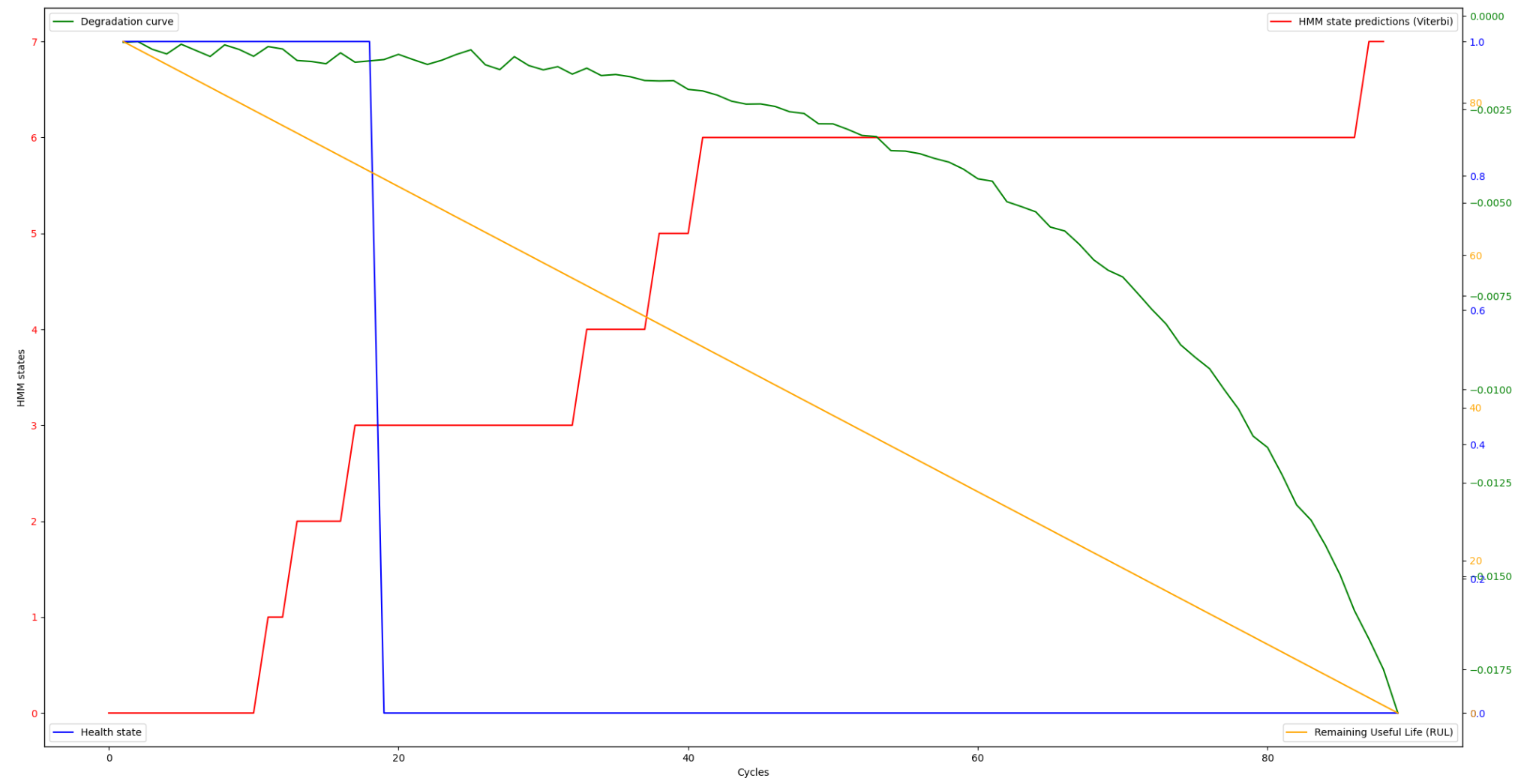}}\quad
      \subfigure[]{\includegraphics[width=0.475\columnwidth]{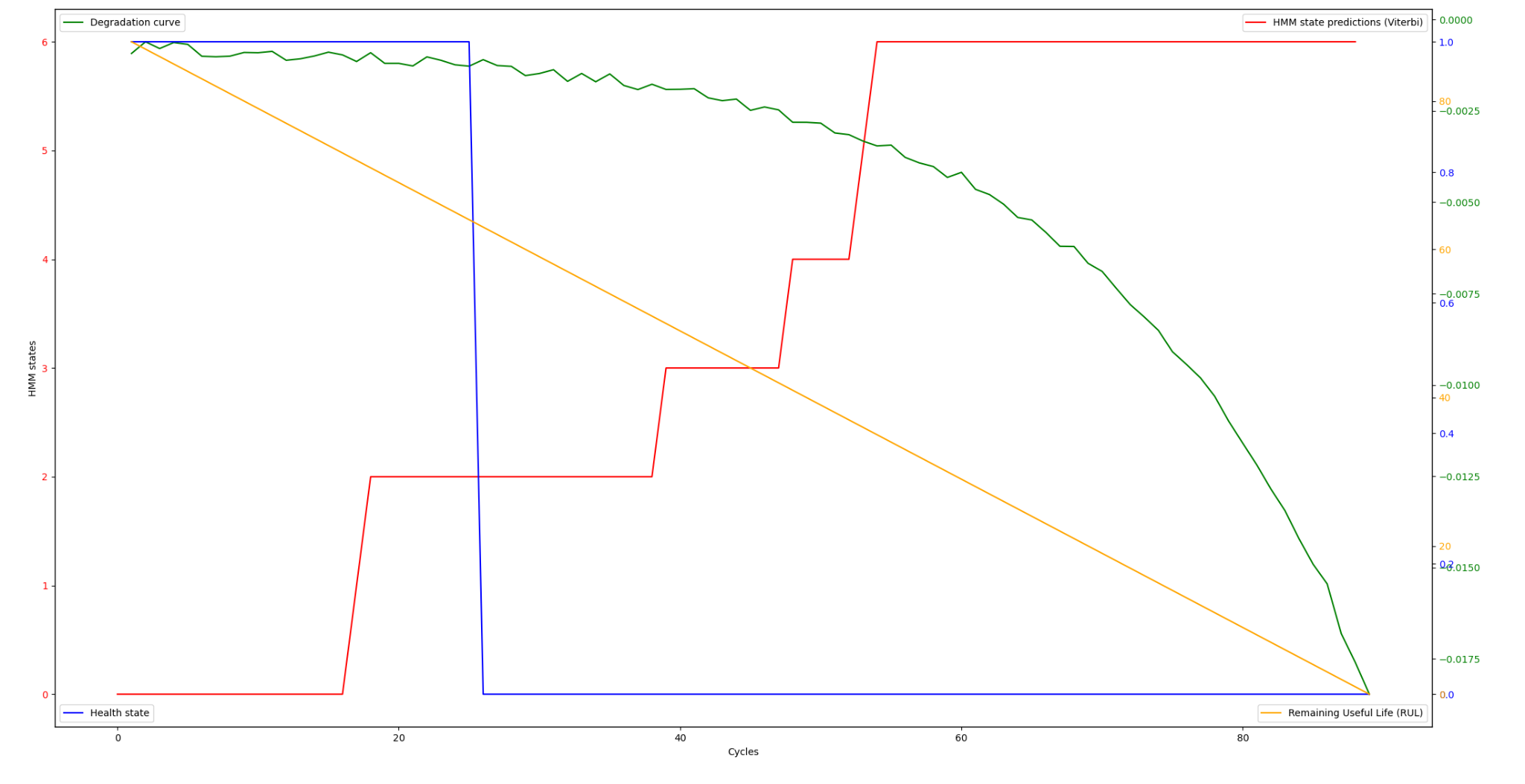}}\quad
      \subfigure[]{\includegraphics[width=0.475\columnwidth]{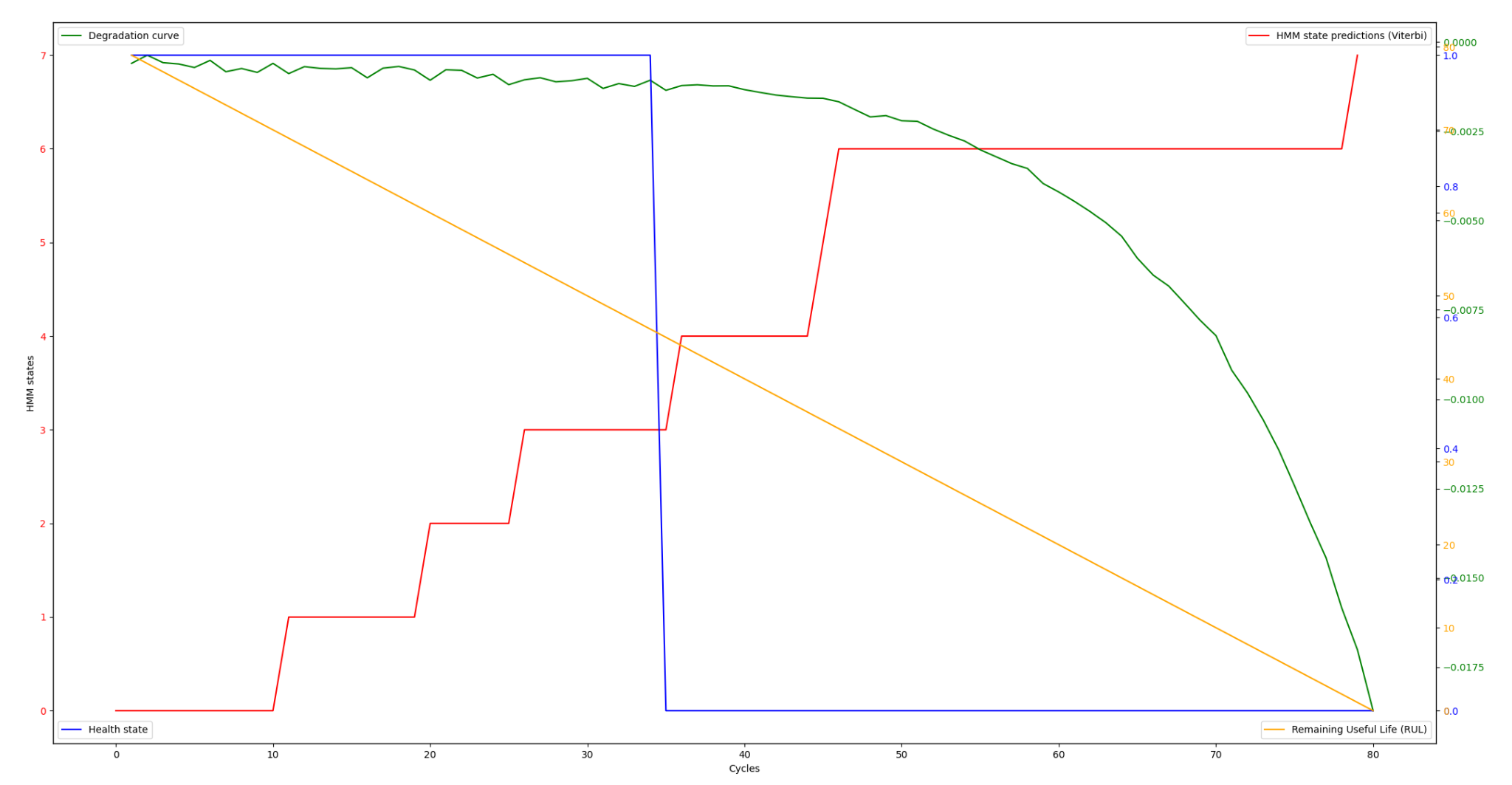}}\quad
      \subfigure[]{\includegraphics[width=0.475\columnwidth]{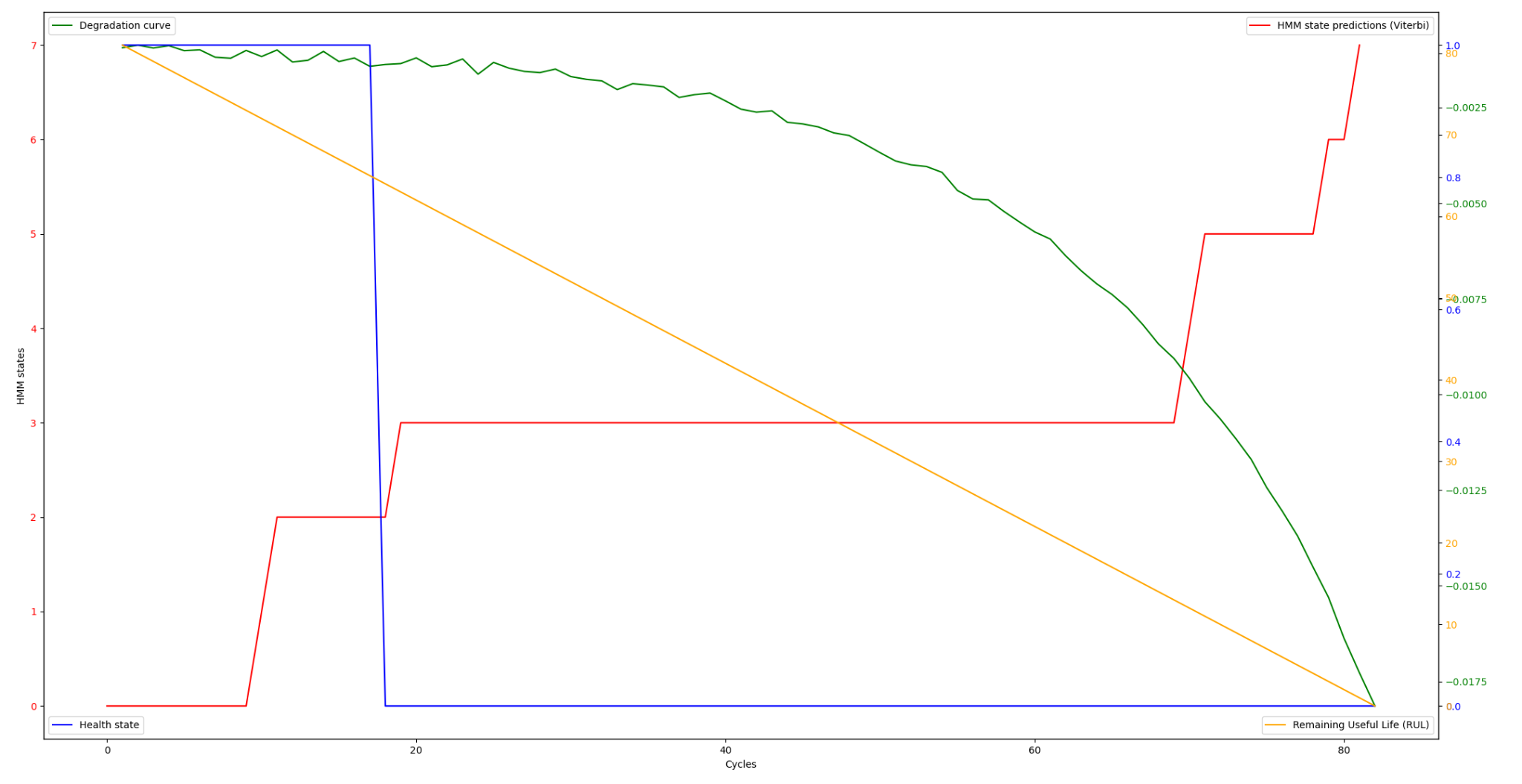}}
      \caption{State decoding and mapping for dataset DS001.}
      \label{fig:state-ds001}
    \end{center} 
\vskip -0.2in
\end{figure}

Apart from the failure event hypothesis, it is necessary to measure the health state of the equipment at different points to generate an alarm for the user when the equipment reaches a critical point of its lifetime. However, the dataset used until now does not contain any such ground truth over which the performance of such interpretability of HMM states can be tested. Therefore, the second version of the dataset \cite{c21} was used to evaluate the state interpretability of HMM throughout the lifetime of the engine and the subset of which is shown in \cref{fig:state-ds001}. The dataset has the ground truth values of engine degradation per equipment, boolean health state value, and remaining useful life. The interpretations were based on the critical points along the equipment degradation curve as shown in  \cref{fig:deg_curve} of \cref{append:fig}. Based on the degradation curve, and the range of HMM states observed during those conditions it was concluded that a relevant set of state hypotheses can be interpreted through the state distribution, as shown in \cref{tab:interpret-ds001}.

\begin{table}[t]
\caption{HMM state interpretability to equipment conditions.}
\label{tab:interpret-ds001}
\vskip 0.15in
\begin{center}
\begin{small}
\begin{sc}
\begin{tabular}{lc}
\hline
\multicolumn{1}{c}{Equipment condition} & HMM states \\ \hline
Normal equipment & 0 - 2 \\
Potential fault point of equipment & 2 - 4 \\
Failure progression & 4 - 6 \\
Fault point of equipment function & 6 - 7 \\
Failure & 7 \\ \hline
\end{tabular}
\end{sc}
\end{small}
\end{center}
\vskip -0.1in
\end{table}

%***************************************************************************************************************************************

\section{Experiment 2: Comparison of SRLA with Baseline and Prior Work}
\label{sec:exp2}
Until now, the dataset used just consisted of 1 operating condition, however, in real-world cases, this is not the case. To adapt to a more general architecture, an Input-Output Hidden Markov Model (IOHMM) is used instead of the HMM. Dataset FD002 is used in this experiment section having 6 operating conditions, which will be used for the comparative evaluation with the prior work, i.e., to the best of our knowledge, the state-of-the-art methodology \cite{sm20} in this particular case. 

%***************************************************************************************************************************************

\subsection{Comparative Evaluation and Results}
\label{sec:exp-res}
As seen in \cref{sec:int-state}, HMM could distribute the states well according to the engine health state, however, the distribution was not fine enough to replace the engine just 1 cycle before the failure, which is the most crucial point for the performance of the replacement policy as designed in this paper. Therefore, DRL is used to refine the granularity after state distribution based on IOHMM, resulting in a hierarchical model. To evaluate the performance, the results are compared with four baseline systems and prior work \cite{sm20}, where the authors have used the Particle Filtering (PF) based-DRL. The authors used 80 engines as the training set and 20 as the test set out of 260 engines. However, the engines were selected randomly, therefore the exact comparison with the average cost of the agent could not be made. Therefore, the ratio of the Ideal Maintenance Cost (IMC) to the average cost of the agent ($\widetilde{Q^{*}}$) was compared in \cref{tab:comparison}. As shown, SRLA outperforms the baseline systems and has a comparative performance with PF + DRL methodology with the added benefits of interpretability. Inference algorithm for SRLA is described in \cref{alg:srla} of \cref{append-algo}.

\begin{table*}[b]
\caption{Comparison of the proposed methodology with baseline systems and \cite{sm20} on dataset FD002.}
\label{tab:comparison}
\vskip 0.15in
\begin{center}
\begin{small}
\begin{sc}
\begin{tabular}{lccc
>{\columncolor[HTML]{D9D9D9}}c
cc
>{\columncolor[HTML]{D9D9D9}}c}
\hline
Methodology & $\widetilde{Q^{*}}$ & IMC & CMC & IMC/$\widetilde{Q^{*}}$ & Failure & Average Remaining cycles & Interpretations \\ \hline
System 1 & 2.10 & 0.64 & 7.02 & 0.30 & 20\% & 5.9 & No \\ 
System 2 & 6.87 & 0.64 & 7.02 & 0.09 & 90\% & 2.6 & No\\ 
System 3 & 7.02 & 0.64 & 7.02 & 0.09 & 100\% & 0.0 & Yes\\ 
System 4 & 0.77 & 0.64 & 7.02 & 0.83 & 0\% & 23.0 & Yes\\ 
PF + DRL {[}15{]} & 2.02 & 1.93 & 20.80 & 0.96 & 0\% & - & No\\
\cellcolor[HTML]{D9EAD3}SRLA & \cellcolor[HTML]{D9EAD3}0.69 & \cellcolor[HTML]{D9EAD3}0.64 & 
\cellcolor[HTML]{D9EAD3}7.02 & \cellcolor[HTML]{D9EAD3}0.94 & \cellcolor[HTML]{D9EAD3}0\% & 
\cellcolor[HTML]{D9EAD3}6.4 & \cellcolor[HTML]{D9EAD3}Yes\\ \hline
\end{tabular}
\end{sc}
\end{small}
\end{center}
\vskip -0.1in
\end{table*}

%***************************************************************************************************************************************

\section*{Conclusion and Future Direction}
\label{conc}

In this paper, a new hierarchical methodology was proposed utilizing the hidden Markov model-based deep reinforcement learning allowing the functionality of interpretability in the stochastic environment along with defining an optimal replacement policy and estimating remaining useful life without supervised annotations. Therefore, such a model can easily be used in industrial cases where the annotation of the fault type is difficult to obtain and the human supervisor in the loop can help define the state distribution according to the event-based analysis. To test the effectiveness of the model, NASA C-MAPSS (turbofan engines) dataset versions 1 and 2 were used. To evaluate the performance, it was compared with baseline models and prior work of Bayesian filtering based-deep reinforcement learning. The results were outperforming and comparable with the added benefits of interpretability with a less complex system model. In the future, the proposed architecture will be used on other open datasets to create a benchmark along with real-world case studies to measure its robustness.

%***************************************************************************************************************************************

%%%%%%%%%%%%%%%%%%%%%%%%%%%%%%%%%%%%%%%%%%%%%%%%%%%%%%%%%%%%%%%%%%%%%%%%%%%%%%%
%%%%%%%%%%%%%%%%%%%%%%%%%%%%%%%%%%%%%%%%%%%%%%%%%%%%%%%%%%%%%%%%%%%%%%%%%%%%%%%
% REFERENCES
%%%%%%%%%%%%%%%%%%%%%%%%%%%%%%%%%%%%%%%%%%%%%%%%%%%%%%%%%%%%%%%%%%%%%%%%%%%%%%%
%%%%%%%%%%%%%%%%%%%%%%%%%%%%%%%%%%%%%%%%%%%%%%%%%%%%%%%%%%%%%%%%%%%%%%%%%%%%%%%

\bibliography{HMM-DRL}

\begin{thebibliography}{34}
\providecommand{\natexlab}[1]{#1}
\providecommand{\url}[1]{\texttt{#1}}
\expandafter\ifx\csname urlstyle\endcsname\relax
  \providecommand{\doi}[1]{doi: #1}\else
  \providecommand{\doi}{doi: \begingroup \urlstyle{rm}\Url}\fi

\bibitem[Bengio \& Frasconi(1995)Bengio and Frasconi]{bf95}
Bengio, Y. and Frasconi, P.
\newblock An input output hmm architecture.
\newblock \emph{Advances in neural information processing systems}, pp.\
  427--434, 1995.

\bibitem[Bertsekas \& Tsitsiklis(1996)Bertsekas and
  Tsitsiklis]{bertsekas1996neuro}
Bertsekas, D.~P. and Tsitsiklis, J.~N.
\newblock \emph{Neuro-dynamic programming}.
\newblock Athena Scientific, 1996.

\bibitem[Brockman et~al.(2016)]{b16}
Brockman, G. et~al.
\newblock Openai gym.
\newblock preprint, arXiv, 2016.

\bibitem[Chao(2021)]{c21}
Chao, A.
\newblock Manuel.
\newblock \emph{et al. "Aircraft Engine Run-to-Failure Dataset under Real
  Flight Conditions for Prognostics and Diagnostics}, 6:\penalty0 1, 2021.

\bibitem[Chen et~al.(2003)]{chen2003bayesian}
Chen, Z. et~al.
\newblock Bayesian filtering: From kalman filters to particle filters, and
  beyond.
\newblock \emph{Statistics}, 182\penalty0 (1):\penalty0 1--69, 2003.

\bibitem[Do et~al.(2015)]{d15}
Do, P. et~al.
\newblock A proactive condition-based maintenance strategy with both perfect
  and imperfect maintenance actions.
\newblock \emph{Reliability Engineering \& System Safety}, 133:\penalty0
  22--32, 2015.

\bibitem[Dulac-Arnold et~al.(2021)Dulac-Arnold, Levine, Mankowitz, Li,
  Paduraru, Gowal, and Hester]{dulac2021challenges}
Dulac-Arnold, G., Levine, N., Mankowitz, D.~J., Li, J., Paduraru, C., Gowal,
  S., and Hester, T.
\newblock Challenges of real-world reinforcement learning: definitions,
  benchmarks and analysis.
\newblock \emph{Machine Learning}, pp.\  1--50, 2021.

\bibitem[Forney(1973)]{forney1973viterbi}
Forney, G.~D.
\newblock The viterbi algorithm.
\newblock \emph{Proceedings of the IEEE}, 61\penalty0 (3):\penalty0 268--278,
  1973.

\bibitem[Giantomassi et~al.(2011)]{g11}
Giantomassi, A. et~al.
\newblock Hidden {M}arkov model for health estimation and prognosis of turbofan
  engines.
\newblock \emph{International Design Engineering Technical Conferences and
  Computers and Information in Engineering Conference}, 5480, 2011.

\bibitem[Griffith et~al.(2013)]{g13}
Griffith, S. et~al.
\newblock \emph{Policy shaping: Integrating human feedback with reinforcement
  learning}.
\newblock Georgia Institute of Technology, 2013.

\bibitem[Hofmann \& Tashman(2020)Hofmann and Tashman]{hofmann2020hidden}
Hofmann, P. and Tashman, Z.
\newblock Hidden markov models and their application for predicting failure
  events.
\newblock In \emph{International Conference on Computational Science}, pp.\
  464--477. Springer, 2020.

\bibitem[Klingelschmidt et~al.(2017)Klingelschmidt, Weber, Simon, Theilliol,
  and Peysson]{7984302}
Klingelschmidt, T., Weber, P., Simon, C., Theilliol, D., and Peysson, F.
\newblock Fault diagnosis and prognosis by using input-output hidden markov
  models applied to a diesel generator.
\newblock In \emph{2017 25th Mediterranean Conference on Control and Automation
  (MED)}, pp.\  1326--1331, 2017.
\newblock \doi{10.1109/MED.2017.7984302}.

\bibitem[Lee et~al.(2015)]{hmmlearn}
Lee, A. et~al.
\newblock hmmlearn.
\newblock \url{https://github.com/hmmlearn/hmmlearn}, 2015.

\bibitem[Lee \& Choi(2021)Lee and Choi]{lee2021attaining}
Lee, J.-H. and Choi, J.
\newblock Attaining interpretability in reinforcement learning via hierarchical
  primitive composition.
\newblock \emph{arXiv preprint arXiv:2110.01833}, 2021.

\bibitem[Lepenioti et~al.(2020)]{l20}
Lepenioti, K. et~al.
\newblock Machine learning for predictive and prescriptive analytics of
  operational data in smart manufacturing.
\newblock In \emph{International Conference on Advanced Information Systems
  Engineering. , Cham}, 2020.

\bibitem[Lyu et~al.(2019)Lyu, Yang, Liu, and Gustafson]{lyu2019sdrl}
Lyu, D., Yang, F., Liu, B., and Gustafson, S.
\newblock Sdrl: interpretable and data-efficient deep reinforcement learning
  leveraging symbolic planning.
\newblock In \emph{Proceedings of the AAAI Conference on Artificial
  Intelligence}, volume~33, pp.\  2970--2977, 2019.

\bibitem[Meng et~al.(2019)Meng, An, Li, and Yang]{meng2019adaptive}
Meng, F., An, A., Li, E., and Yang, S.
\newblock Adaptive event-based reinforcement learning control.
\newblock In \emph{2019 Chinese Control And Decision Conference (CCDC)}, pp.\
  3471--3476. IEEE, 2019.

\bibitem[Ong et~al.(2020)Ong, Niyato, and Yuen]{ony20}
Ong, K. S.~H., Niyato, D., and Yuen, C.
\newblock Predictive maintenance for edge-based sensor networks: A deep
  reinforcement learning approach.
\newblock In \emph{2020 IEEE 6th World Forum on Internet of Things (WF-IoT)},
  pp.\  1--6. IEEE, 2020.

\bibitem[Panzer \& Bender(2021)Panzer and Bender]{pb21}
Panzer, M. and Bender, B.
\newblock Deep reinforcement learning in production systems: a systematic
  literature review.
\newblock \emph{International Journal of Production Research}, pp.\  1--26,
  2021.

\bibitem[Parra-Ullauri et~al.(2021)Parra-Ullauri, Garc{\'\i}a-Dom{\'\i}nguez,
  Bencomo, Zheng, Zhen, Boubeta-Puig, Ortiz, and Yang]{parra2021event}
Parra-Ullauri, J.~M., Garc{\'\i}a-Dom{\'\i}nguez, A., Bencomo, N., Zheng, C.,
  Zhen, C., Boubeta-Puig, J., Ortiz, G., and Yang, S.
\newblock Event-driven temporal models for explanations-etemox: explaining
  reinforcement learning.
\newblock \emph{Software and Systems Modeling}, pp.\  1--23, 2021.

\bibitem[Pateria et~al.(2021)Pateria, Subagdja, Tan, and
  Quek]{pateria2021hierarchical}
Pateria, S., Subagdja, B., Tan, A.-h., and Quek, C.
\newblock Hierarchical reinforcement learning: A comprehensive survey.
\newblock \emph{ACM Computing Surveys (CSUR)}, 54\penalty0 (5):\penalty0 1--35,
  2021.

\bibitem[Rabiner \& Juang(1986)Rabiner and Juang]{1165342}
Rabiner, L. and Juang, B.
\newblock An introduction to hidden markov models.
\newblock \emph{IEEE ASSP Magazine}, 3\penalty0 (1):\penalty0 4--16, 1986.
\newblock \doi{10.1109/MASSP.1986.1165342}.

\bibitem[Rabiner(1989)]{rabiner1989tutorial}
Rabiner, L.~R.
\newblock A tutorial on hidden markov models and selected applications in
  speech recognition.
\newblock \emph{Proceedings of the IEEE}, 77\penalty0 (2):\penalty0 257--286,
  1989.

\bibitem[Saxena \& Goebel(2008)Saxena and Goebel]{sg08}
Saxena, A. and Goebel, K.
\newblock Turbofan engine degradation simulation data set.
\newblock \emph{NASA Ames Prognostics Data Repository :}, pp.\  878--887, 2008.

\bibitem[Shahin et~al.(2019)Shahin, Simon, and Weber]{shahin2019estimating}
Shahin, K.~I., Simon, C., and Weber, P.
\newblock Estimating iohmm parameters to compute remaining useful life of
  system.
\newblock In \emph{Proceedings of the 29th European Safety and Reliability
  Conference, Hannover, Germany}, pp.\  22--26, 2019.

\bibitem[Sikorska et~al.(2011)Sikorska, Hodkiewicz, and
  Ma]{sikorska2011prognostic}
Sikorska, J., Hodkiewicz, M., and Ma, L.
\newblock Prognostic modelling options for remaining useful life estimation by
  industry.
\newblock \emph{Mechanical systems and signal processing}, 25\penalty0
  (5):\penalty0 1803--1836, 2011.

\bibitem[Skordilis \& Moghaddass(2020)Skordilis and Moghaddass]{sm20}
Skordilis, E. and Moghaddass, R.
\newblock A deep reinforcement learning approach for real-time sensor-driven
  decision making and predictive analytics.
\newblock \emph{Computers \& Industrial Engineering}, 147, 2020.

\bibitem[Sutton \& Barto(2018)Sutton and Barto]{sb18}
Sutton, R.~S. and Barto, A.~G.
\newblock \emph{Reinforcement learning: An introduction}.
\newblock MIT press, 2018.

\bibitem[Van~Hasselt et~al.(2016)Van~Hasselt, Guez, and Silver]{vgs16}
Van~Hasselt, H., Guez, A., and Silver, D.
\newblock Deep reinforcement learning with double q-learning.
\newblock In \emph{Proceedings of the AAAI conference on artificial
  intelligence}. Vol. 30. No. 1, 2016.

\bibitem[Xu \& Fekri(2021)Xu and Fekri]{xu2021interpretable}
Xu, D. and Fekri, F.
\newblock Interpretable model-based hierarchical reinforcement learning using
  inductive logic programming.
\newblock \emph{arXiv preprint arXiv:2106.11417}, 2021.

\bibitem[Yin \& Silva(2017)Yin and Silva]{iohmm}
Yin, M. and Silva, T.
\newblock Iohmm.
\newblock \url{https://github.com/Mogeng/IOHMM}, 2017.

\bibitem[Yoon et~al.(2019{\natexlab{a}})Yoon, Lee, and
  Hovakimyan]{yoon2019hidden}
Yoon, H.-J., Lee, D., and Hovakimyan, N.
\newblock Hidden markov model estimation-based q-learning for partially
  observable markov decision process.
\newblock In \emph{2019 American Control Conference (ACC)}, pp.\  2366--2371.
  IEEE, 2019{\natexlab{a}}.

\bibitem[Yoon et~al.(2019{\natexlab{b}})Yoon, Lee, and
  Hovakimyan]{yoon_lee_hovakimyan_2019}
Yoon, H.-J., Lee, D., and Hovakimyan, N.
\newblock Hidden markov model estimation-based q-learning for partially
  observable markov decision process.
\newblock \emph{2019 American Control Conference (ACC)}, 2019{\natexlab{b}}.
\newblock \doi{10.23919/acc.2019.8814849}.

\bibitem[Yu(2018)]{yu2018towards}
Yu, Y.
\newblock Towards sample efficient reinforcement learning.
\newblock In \emph{IJCAI}, pp.\  5739--5743, 2018.

\end{thebibliography}
\bibliographystyle{icml2022}

%%%%%%%%%%%%%%%%%%%%%%%%%%%%%%%%%%%%%%%%%%%%%%%%%%%%%%%%%%%%%%%%%%%%%%%%%%%%%%%
%%%%%%%%%%%%%%%%%%%%%%%%%%%%%%%%%%%%%%%%%%%%%%%%%%%%%%%%%%%%%%%%%%%%%%%%%%%%%%%
% APPENDIX
%%%%%%%%%%%%%%%%%%%%%%%%%%%%%%%%%%%%%%%%%%%%%%%%%%%%%%%%%%%%%%%%%%%%%%%%%%%%%%%
%%%%%%%%%%%%%%%%%%%%%%%%%%%%%%%%%%%%%%%%%%%%%%%%%%%%%%%%%%%%%%%%%%%%%%%%%%%%%%%
\newpage
\appendix
\onecolumn

%%%%%%%%%%%%%%%%%%%%%%%%%%%%%%%%%%%%%%%%%%%%%%%%%%%%%%%%%%%%%%%%%%%%%%%%%%%%%%%
\renewcommand\thealgorithm{\thesection.\arabic{algorithm}}    
\section{\emph{Algorithms and Training Parameters}}
\label{append-algo}
\setcounter{algorithm}{0}

%***************************************************************************************************************************************

\begin{algorithm}[ht]
   \caption{Environment Modeling}
   \label{alg:env_mod}
\begin{algorithmic}
   \STATE {\bfseries Input:} {\newline 
                              $S_t = {s_0,...,s_t,...,s_T}$: state space \newline 
                              $A_t = {a_0,...,a_n}$: action space \newline 
                              $R_t(s_t, a_t)$: reward given current state and chosen action}
   \REPEAT
   \STATE step in environment and sample observed state and reward
   \IF{$hold$}
       \IF{$\textrm{equipment has reached the failure state}$}
            \STATE reward of failure   
            \STATE end of episode
            \STATE replace to new different equipment and observe $S_0^{m+1}$
       \ELSE
            \STATE reward of hold   
            \STATE increasing the age of the equipment by one step  
            \STATE observed next state: $S_{t+1}$
        \ENDIF
    \ELSIF{$replace$}
        \IF{$\textrm{equipment has reached the failure state}$}
            \STATE reward of failure   
        \ELSE
            \STATE reward of replacement   
        \ENDIF
            \STATE end of episode
            \STATE replace to new different equipment and observe $S_0^{m+1}$
   \ENDIF
   \STATE {\bfseries Output:} $S_{t+1}$ : next state, $R_t$:  reward, end of episode 
   \UNTIL{$\textrm{all states have been observed}$}
\end{algorithmic}
\end{algorithm}

%***************************************************************************************************************************************

\begin{algorithm}[H]
   \caption{Approximate Reinforcement Learning}
   \label{alg:DRL}
\begin{algorithmic}
   \STATE {\bfseries Input:} {\newline 
                              $T_{max}$: epochs \newline  
                              $\epsilon_0$: initial exploration parameter \newline 
                              $\gamma$: discount factor \newline  
                              $\alpha$: learning rate \newline 
                              $S_t = {s_0,...,s_t,...,s_T}$: state space \newline 
                              $A_t = {a_0,...,a_n}$: action space \newline 
                              $R_t(s_t, a_t)$: reward given current state and chosen action}
   \FOR{$t = 1$ {\bfseries to} $T_{max}$}
   
   \IF{$training$}
   \STATE $\epsilon = \epsilon_0$
   \ELSE
   \STATE $\epsilon = 0$
   \ENDIF
   
   \IF{$exploration < \epsilon$}
   \STATE choose action randomly
   \ELSE
   \STATE Choose action with maximum q-value
   \ENDIF
   
   \STATE find $S_{t+1}$ and $R_t$ given the chosen action
   \STATE approximate $Q$ for actions in current states $\hat{Q}(s_t, a_t)$
   \STATE sum approximated $Q$ for chosen actions $\mathbb{E}\{\hat{Q}(s_t, a_t)\}$
   \STATE approximate $Q$ for actions in next state $\hat{Q}(s_{t+1}|s_t, a_t)$	
   \STATE compute $Q\mbox{*}(s_{t+1}|s_t, a_t) = max(\hat{Q}(s_{t+1}|s_t, a_t))$
   \STATE $Q_{target} = R_t + \gamma(Q\mbox{*}(s_{t+1}|s_t, a_t))$
   \STATE update $\hat{Q}$ by MSE between target and previous value;
   \STATE $\hat{Q}(s_t, a_t) \rightarrow \hat{Q}(s_t, a_t) + \alpha(\frac{1}{n}\Sigma(Q_{target} - \hat{Q}(s_t, a_t)))^2$
   \STATE $\textrm{total reward} = \Sigma(R_t)$
   \STATE $S_t  = S_{t+1}$
   \STATE decay epsilon by defined percentage
   
   \IF{$\textrm{last state \textbf{or} end of episode}$}
   \STATE $Q_{target} = R_t(s_t, a_t)$
   \STATE break the loop
   \ENDIF 

   \ENDFOR
   \STATE {\bfseries Output:} $\hat{Q\mbox{*}}(s_t, a_t)$
\end{algorithmic}
\end{algorithm}

%***************************************************************************************************************************************

\begin{algorithm}[H]
   \caption{Feature Importance}
   \label{alg:feat-imp}
\begin{algorithmic}
   \STATE {\bfseries Input:} {\newline 
                              Viterbi state predictions as target classes: $\hat{S}$ \newline
                              Normalized sensor readings as features: $X$}
   \REPEAT
   \STATE Fit features with the classes
   \STATE Extract the relevance of features corresponding to every class
   \STATE {\bfseries Output:} Feature relevance
   \UNTIL{$\textrm{all features are evaluated for concerned states}$}
\end{algorithmic}
\end{algorithm}

%***************************************************************************************************************************************

\begin{algorithm}[H]
   \caption{RUL Estimation}
   \label{alg:rul_est}
\begin{algorithmic}
   \STATE {\bfseries Input:} {\newline 
                              $Y$: observation sequence till cycle $‘t’$ \newline
                              $A$: Transition probability matrix \newline
                              $B$: Emission probabilities}
   \FOR{$every \: cycle$}
       \STATE useful cycles = 0
       \WHILE{$predicted \: state \: is \: not \: in \: failure \: state$}
            \FOR{$100 \: iterations$}
                \STATE Predict state sequence using Viterbi algorithm
                \STATE Select most probable next state
                \STATE Sample sensor observations of predicted state
                \STATE Append predicted state to state sequence
                \STATE Append sampled observation to sequence
                \STATE Add 1 to ‘useful cycles’
            \ENDFOR 
            \STATE RUL = Average of the useful cycles
        \ENDWHILE
    \ENDFOR  
    \STATE RUL for each point = list of RULs
    \STATE {\bfseries Output:} RUL prediction
\end{algorithmic}
\end{algorithm}

%***************************************************************************************************************************************

\begin{algorithm}[H]
   \caption{Specialized Reinforcement Learning Agent (SRLA)}
   \label{alg:srla}
\begin{algorithmic}
\STATE {\bfseries \emph{STEP I:}} IOHMM Training
   \STATE {\bfseries Input:} {\newline 
                              $n$: number of hidden states \newline  
                              $Y$: output sequences \newline 
                              $U$: input seauences} 
   \STATE {\bfseries Output:} $\lambda$: model parameters (initial, transition, and emission probability) \newline 

\STATE {\bfseries \emph{STEP II:}} Viterbi Algorithm (IOHMM inference)
   \STATE {\bfseries Input:} {$\lambda$, $U$, $Y$} 
   \STATE {\bfseries Output:} $\delta_{t}(i)  =  \max _{x_{1}, \cdots, x_{t-1}} P\left[x_{1} \cdots x_{t}=i, u_{1} \cdots u_{t}, y_{1} \cdots y_{t} \mid \lambda\right]$ \newline 

\STATE {\bfseries \emph{STEP III:}} DRL Training
   \STATE {\bfseries Input:} {\newline 
                              $\delta_{s}$: specific event (such as failure) \newline  
                              $S_t$: $u_t$ + $y_t$} 
   \STATE \cref{alg:env_mod}
   \STATE \cref{alg:DRL}
   \STATE {\bfseries Output:} $\hat{Q\mbox{*}}(S_t, A_t)$ \newline 

\STATE {\bfseries \emph{STEP IV:}} SRLA Inference
   \STATE {\bfseries Input:} {$\lambda$, $\hat{Q\mbox{*}}(S_t, A_t)$, $S_t$: $(U_t, Y_t)$}
   \STATE Step II, \cref{alg:feat-imp}, and \cref{alg:rul_est} for interpretations
   \STATE $\delta \rightarrow$ Specialized state ($X_s$) $\rightarrow U_s, Y_s$
   \IF{$S_t$ in $X_s$}
        \STATE $\hat{Q\mbox{*}}(s_t, a_t)$
        \STATE \cref{alg:env_mod}
    \ELSE 
    \STATE $a_t =$ do nothing (hold)
    \ENDIF 
   \STATE {\bfseries Output:} $\hat{Q\mbox{*}}(\delta_t, s_t, a_t)$
\end{algorithmic}
\end{algorithm}

\subsection{Training Parameters}
\label{app-sec:train_param}
The summary of the DL framework within the RL architectures are as follows: (a) Deep Neural Network (DNN) consisting of a total of ~37,000 training parameters and fully-connected (dense) layers with 2 hidden layers having 128 and 256 neurons, respectively, with ReLU activation. (b)  Recurrent Neural Network (RNN) consists of a total of ~468,000 training parameters and fully connected (LSTM) layers with 2 hidden layers having 128 and 256 neurons, respectively. The output layer consists of the number of actions the agent can decide for decision-making with linear activation. The parameters of the DRL agent are as follows: discount rate = 0.95, learning rate = 1e-4, and the epsilon decay rate = 0.99 is selected with the initial epsilon = 0.5. 

%%%%%%%%%%%%%%%%%%%%%%%%%%%%%%%%%%%%%%%%%%%%%%%%%%%%%%%%%%%%%%%%%%%%%%%%%%%%%%%
\renewcommand\thefigure{\thesection.\arabic{figure}}    
\pagebreak
\section{\emph{Figures}}
\label{append:fig}
\setcounter{figure}{0} 

\begin{figure}[!htb]
\vskip 0.2in
\begin{center}
\centerline{\includegraphics[width=0.55\columnwidth]{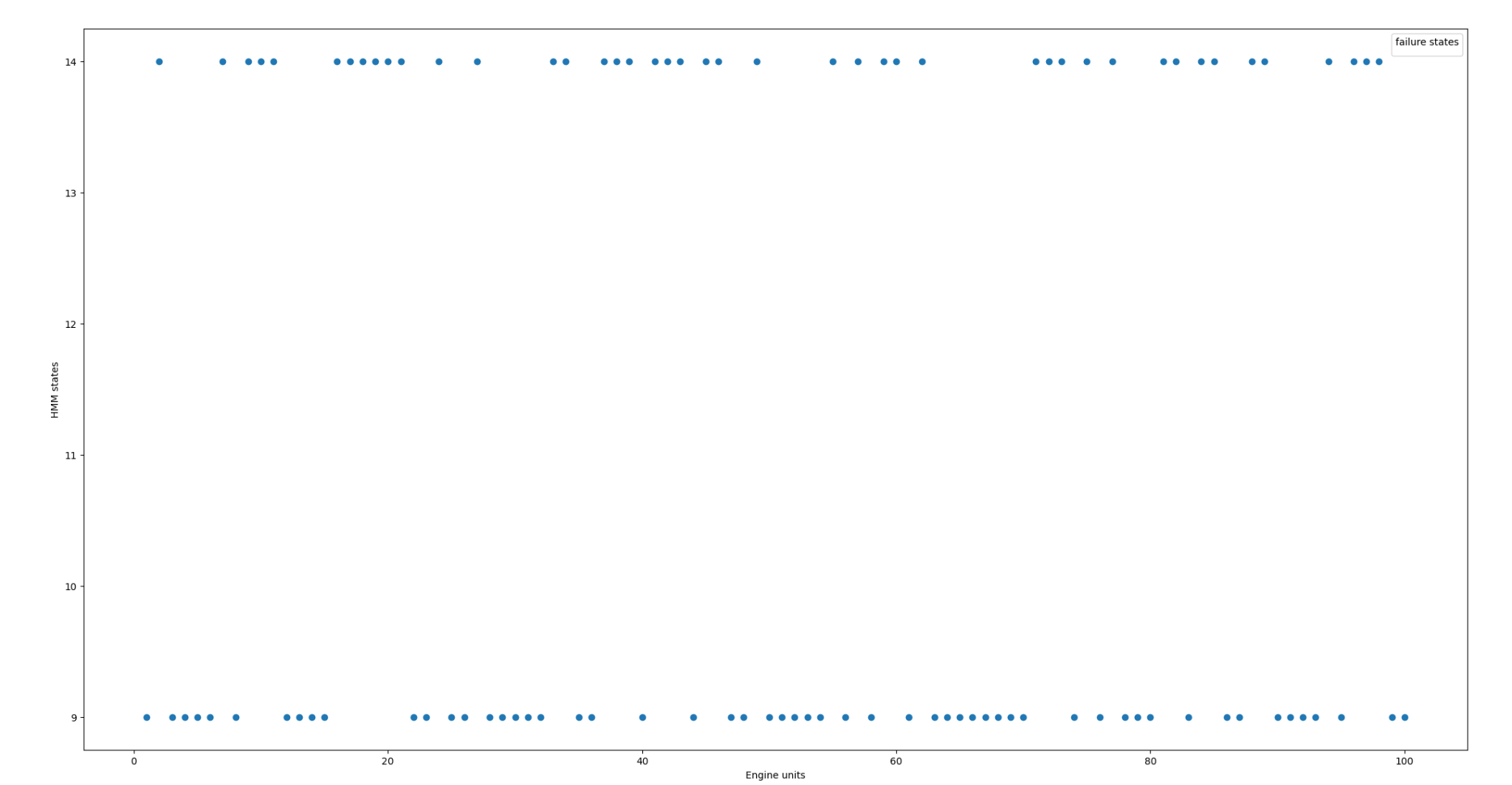}}
\caption{States decoding and mapping for dataset FD003.}
\label{fig:state_decoding-fd003}
\end{center}
\vskip -0.2in
\end{figure}

\begin{figure}[!htb]
\vskip 0.2in
\begin{center}
\centerline{\includegraphics[width=0.55\columnwidth]{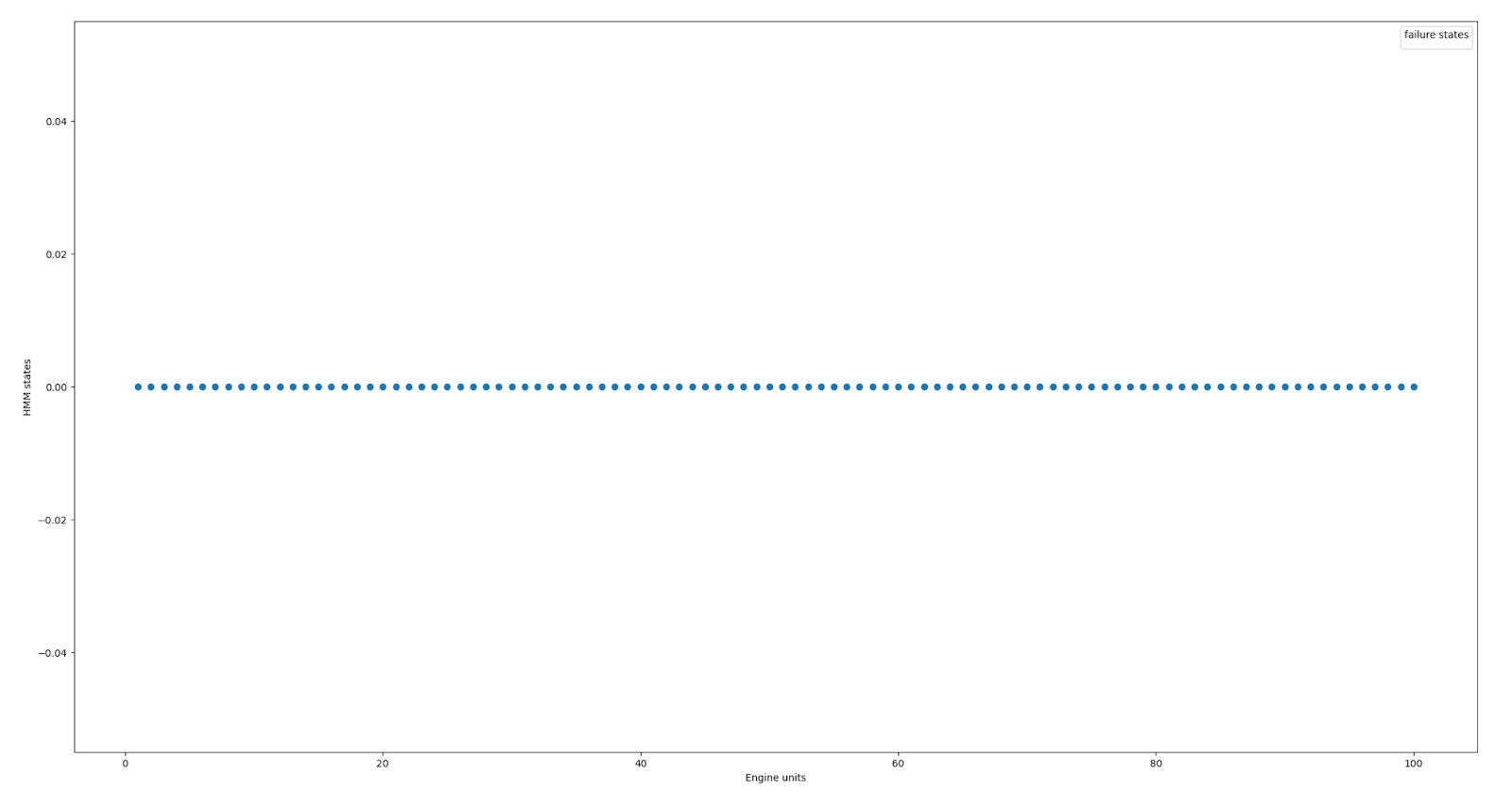}}
\caption{States decoding and mapping for dataset FD001.}
\label{fig:state_decoding-fd001}
\end{center}
\vskip -0.2in
\end{figure}

\begin{figure}[!htb]
\vskip 0.2in
\begin{center}
\centerline{\includegraphics[width=0.55\columnwidth]{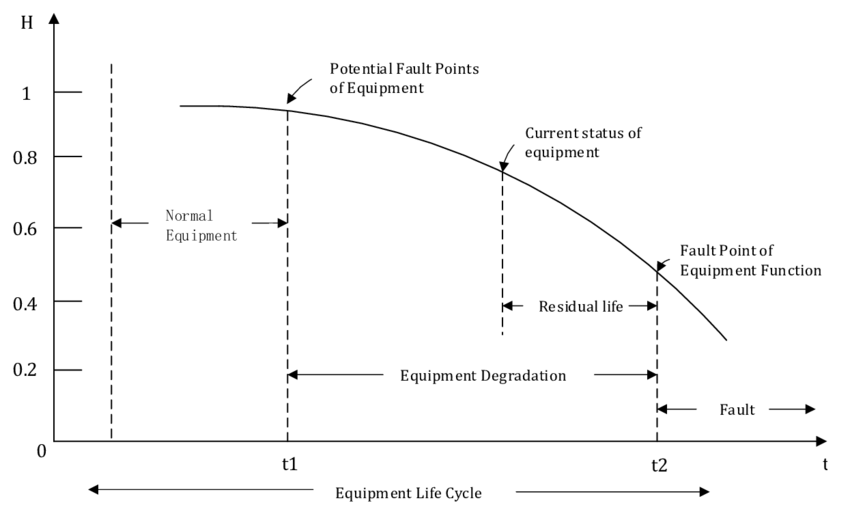}}
\caption{Health degradation curve of equipment.}
\label{fig:deg_curve}
\end{center}
\vskip -0.2in
\end{figure}

%%%%%%%%%%%%%%%%%%%%%%%%%%%%%%%%%%%%%%%%%%%%%%%%%%%%%%%%%%%%%%%%%%%%%%%%%%%%%%%

\end{document}